%% file: DNN_training.tex
\pdfoutput=1
\documentclass{article}

 \usepackage[preprint]{nips_2018}

\input{Packages.tex}
\input{Macros.tex}

\title{A Unified Framework for Training Neural Networks}

 \author{
   Hadi~Ghauch$^\star$ \And Hossein Shokri-Ghadikolaei$^\star$ \And Carlo Fischione$^\star$ \And Mikael Skoglund
   \thanks{School of EECS, KTH Royal Institute of Technology, Stockholm, 10044
   \newline Email: \texttt{ghauch@kth.se, hshokri@kth.se, carlofi@kth.se, skoglund@kth.se}  } }

\begin{document}

\maketitle

\begin{abstract}
The lack of mathematical tractability of Deep Neural Networks (DNNs) has hindered progress towards having a unified convergence analysis of training algorithms, in the general setting.
We propose a unified optimization framework for training different types of DNNs, and establish its convergence for arbitrary loss, activation, and regularization functions, assumed to be smooth. We show that framework generalizes well-known first- and second-order training methods, and thus allows us to show the convergence of these methods for various DNN architectures and learning tasks, as a special case of our approach. We discuss some of its applications in training various DNN architectures (e.g., feed-forward, convolutional, linear networks), to regression and classification tasks.
\end{abstract}

\section{Introduction} \label{sec:Intro}

{Deep neural networks (DNNs)} have been successfully applied to many machine learning and statistical inference problems including speech recognition~\citep{hinton2012deep}, natural language processing~\citep{collobert2008unified}, and image classification~\citep{krizhevsky2012imagenet}. Although the ability of a DNN to approximate any smooth function was known since the 80's~\citep{cybenko1989approximation,hornik1989multilayer,barron1993universal}, computational approaches to the find optimal weights in the network usually lack theoretical guarantees. The main reason that the involved optimization problems are non-convex~\citep{glorot2010understanding,zhang2017learnability} and NP-hard~\citep{blum1989training}. In practice, however, various local search methods, e.g., stochastic gradient descent and \emph{back propagation~(BP)}~\citep{LeCun_EfficientBP_98} show excellent training performance~\citep{kingma2014adam}, though their success is not very well understood.

\textbf{Related Work:}~\citet{mei2016landscape} and~\citet{hazan2015beyond} studied the convergence of the local search algorithms to train a neural network with no hidden layer, where the associated optimization problem has a single minimizer.~\citet{tian2017analytical} and~\citet{zhong2017recovery} consider a two-layer neural network with Gaussian inputs under specific assumptions on the weights and number of hidden layers, thus making the resulting loss function convex in a small neighborhood around the global solution. 
Deep linear networks (i.e., DNNs with identity activation functions) have also gained much attention lately, due to their analytical tractability~\citep{Yun_GlobOptDNN_17}. Moreover, increasing the number of hidden layers in such networks does not result in 'bad' local optima~\citep{Lu_DLN_17}.  
Recently,\citet{Yun_GlobOptDNN_17} derived global optimality conditions for deep linear networks. Other approaches for training DNNs, using ADMM,  were explored but their convergence remains unknown ~\citep{Taylor_DNNTrainADMM_16}.~\citet{Lorenzo_DistML_SCA_16} studied distributed training via successive convex approximations of the loss (in a single-layer neural network), focusing on communication-related aspects.

While these works provide promising initial analytical results for training DNNs, they occupy a rather narrow slice in the large DNN training literature. As these methods/results are applicable to specific scenarios, we still lack a  comprehensive framework for analyzing the convergence of these methods (as opposed to a mere test and try approach), in a general DNN training setup. 
This issue is of paramount importance. 

Our focus in this work is to develop a general framework for DNN training, using variants of \emph{Block-Coordinate Descent (BCD)} methods, whereby we split-up a non-convex (coupled) optimization problem into a series of subproblems (one per block of variables/coordinates). However, strong convexity of each subproblem is a necessary condition to ensure convergence~\citep{Razaviyayn_BCD_12} - an assumption that rarely holds in general DNN training problems.
We remedy the problem by leveraging extensions of BCD methods, namely \emph{Block-Successive Upperbound Minimization (BSUM)}, where the non-convex subproblem is approximated with a carefully constructed convex upperbound~\citep{Razaviyayn_BCD_12}. We thus devise a \emph{general training method}, applicable to a wide range of DNNs and learning tasks.
Similar ideas on the use of \emph{surrogate functions} have appeared in the BCD literature~\citep{Mairal_FoSF_13}.
A BCD training algorithm was recently proposed by~\citet{Zhang_TrainDDN_BCD_17}, assuming convex loss functions,
Tikhonov regularization and ReLU activation function. In that sense, our algorithm \emph{generalizes} these results as it makes minimal assumptions on the loss, activation, and regularization functions.


\textbf{Contributions:}
We adopt in this paper a \emph{general DNN training setting}, which encompasses \emph{several architectures} (feed-forward, convolutional, linear, etc.), and assumes \emph{arbitrary} (yet smooth) loss and activation functions. 
We propose a new optimization framework offering \emph{provable convergence} guarantees, and establish convergence of several algorithms under it, in this general setting. 
We show that our algorithm can recover many existing training algorithms as special cases, and give sufficient conditions for convergence of these approaches, for general training problems.
Additionally, we identity choices of loss and activations functions, which result in convex subproblems where standard BCD methods may be used.
We also highlight applications of the proposed framework for \emph{general learning tasks} (regression,  classification), and \emph{various DNN architectures }(feed-forward, convolutional, linear).
We underline that this initial investigation is primarily centered around algorithms resulting from first- and second-order upperbounds, under our proposed framework.
Future investigations of other upperbounds, and the resulting algorithms are quite promising.

We present the model and problem formulation in Section~\ref{sec:sysmod}, the proposed algorithm in Section~\ref{sec:prop_appr}, explore several choices for the upperbound in~Section~\ref{sec:UBchoice}, and discuss some applications in training DNNs~in Section~\ref{sec:application}. Due to space limitations, we have moved all the proofs and extra discussions to the appendix.



\textbf{Notation:} We use bold upper-case letters to denote matrices, bold lower-case letters to denote vectors, and calligraphic letters to denote sets. For a given matrix $\bA$, $\Vert \pmb{A} \Vert_F$ its Frobenius norm, and $\bA^T$ its transpose.
$\bI_n$ denotes the $n \times n$ identity matrix, $\IB$ the binary set, $\IR^{n}$ is the $n$-dimensional Euclidean space, and $\lrb{n}$ the set of natural numbers from $1$ to $n$.

\section{System Model} \label{sec:sysmod}

We consider a \emph{batch training} method with $N$ samples, $\calT = \lrb{(\bx_n , \by_n )}_{n=1}^N  $, where $\bx_n \in \IR^{d_0}$ and $\by_n \in \IR^{d_J} $ denote the input and output of sample $n \in \lrb{N} $. We let $ \bY = [ \by_1, \cdots , \by_N ] $ , and $ \bX = [ \bx_1, \cdots , \bx_N ] $.
We model the input-output mapping by a DNN with $J$ layers as follows:
\begin{align}
\bZ_0 \rightarrow \boxed{\bW_1 } \rightarrow \boxed{\bSig_1 } \rightarrow \underbrace{\bZ_1}_{ \bSig_1(\bW_1 \bZ_0 )} \rightarrow \boxed{\bW_2  } \rightarrow \boxed{\bSig_2 }
\rightarrow \underbrace{\bZ_2}_{ \bSig_2(\bW_2 \bZ_1)} \rightarrow \cdots \underbrace{\bZ_J}_{\bSig_J(\bW_J \bZ_{J-1} )  }
\end{align}
Note that $\bW_j \in \calW_j \subseteq \IR^{d_j \times d_{j-1}} $ is the matrix of weights connecting layer $j$ to the previous one (dropping the bias without loss in generality), where $ \calW_j $ denotes a closed convex set of feasible weights for $\bW_j$.
Moreover,  $\bZ_j \in \IR^{d_j \times N } $ is the batch output of layer $j$, $\bZ_0 \in \IR^{d_0 \times N} $ denotes the batch input of the network, and $\bZ_J \in \IR^{d_J \times N } $ is its output. $\bSig_j (\bU): \IR^{d_j \times N } \rightarrow \IR^{d_j \times N}  $ is the matrix of \emph{non-linear activation} functions for layer $j$. Thus, the operation of layer $j$ is modeled as,  $\bZ_{j} = \bSig_j( \bW_j \bZ_{j-1} ) , ~j \in \lrb{J}$. Moreover, we denote by $\bH(\bZ_0): \IR^{d_0 \times N } \rightarrow \IR^{d_J \times N }$ the (matrix-valued) \emph{mapping} representing the entire network,
$\bH(\bZ_0) \triangleq ~\bSig_J \left( \bW_J \cdots \bSig_2(\bW_2 ~\bSig_1( \bW_1 \bZ_0 )) \right).$

\subsection{Problem Formulation}
We adopt a general DNN training setting (see Section~\ref{sec:Intro}), by minimizing a \emph{regularized loss} function $f$, which comprises of a \emph{loss} function $\ell$ (i.e., a distance measure between inputs and outputs) and a \emph{regularization} function $r_j$ for layer $j$ (to prevent overfitting). Formally, 
\begin{align} \label{eq:loss_gen}
 f( \bbW )= \ell \left( \overline{\bW} , \bH(\bX), \bY \right) + \sum_{j=1}^J r_j(\bW_j)   &\triangleq \ell( \bbW )  + \sum_{j=1}^J r_j(\bW_j)~,
\end{align}
where $\bbW = \lrb{\bW_1, \cdots, \bW_J} $ is the set of all weights, and
$r_j (\bW_j) : \IR^{d_j \times d_{j-1}} \rightarrow \IR  $. 
\vspace{-.2cm}
\begin{assumption}. \label{ass:loss} \rm
The loss function, $\ell(\bW_1, \cdots, \bW_J ): \IR^{d_1 \times d_{0}}  \times \cdots \times \IR^{d_J \times d_{J-1}} \rightarrow \IR $ is differentiable and Lipschitz continuous.
The regularization function, $r_j (\bW_j) : \IR^{d_j \times d_{j-1}} \rightarrow \IR  $ is assumed to be  differentiable and strongly convex, with constant $\lambda_j > 0$.
Extensions to non-smooth regularization are discussed Appendix~\ref{app:noncovex_reg}.
\end{assumption}
\vspace{-.4cm}
\begin{assumption}. \label{ass:activation} \rm
$\bSig_j (\bU): \IR^{d_j \times N } \rightarrow \IR^{d_j \times N}  $ is the non-linear activation function for layer $j$ (in matrix form). We assume $\bSig_j (\bU)$ to be differentiable in $\bU$ (though not necessarily convex) and satisfying the bounded-input bounded-output property. 
\end{assumption}
\vspace{-.3cm}
Note that such assumptions are quite prevalent in DNN training literature.
With that in mind, we find the optimal weights of the networks by formulating an {empirical risk minimization} problem as:
\begin{align}
(P)
\begin{cases}
\underset{ \overline{\pmb{W}} \in \calW }{\min}~ f( \bbW) = \ell(\bbW) +  \sum_{j=1}^J r_j(\bW_j) ,
\end{cases}
\end{align}
where $\calW \triangleq \Pi_{j=1}^J \calW_j $, and each $\calW_j$ is a closed convex set.
As seen from the above model, this formulation is quite generic: $\ell$ and $r_j$ can model several DDN regression and classification problems, and various architectures (modeled by $\calW$).
In general, $f( \bbW )$  is not jointly convex in the weights due to the apparent coupling among the variables, and the non-linear activation functions. Naturally, the coupling also motivates a BCD approach.



\section{Proposed Approach} \label{sec:prop_appr}
BCD methods operate by optimizing a single block of variables/coordinates during each iteration, assuming all remaining blocks are fixed.
At iteration $k$, blocks $\bW_1, ..., \bW_{j-1}, \bW_{j+1}, \cdots, \bW_J$ are fixed, and block $\bW_j$ is updated to minimize:
\begin{align} \label{eq:f_jdef}
 f_j(  \bW_j ; \bW_{-j}^{(k)} ) \triangleq  \ell( \bW_1^{(k)}, \cdots, \bW_{j-1}^{(k)}, \bW_j , \bW_{j+1}^{(k)}, \cdots, \bW_J^{(k)} ) + r_j(\bW_j) ~.
\end{align}
In the above, superscript ${(k)}$ is the iteration number, and $\bW_{-j}^{(k)} = \lrb{ \bW_1^{(k)}, \cdots, \bW_{j-1}^{(k)}, \bW_{j+1}^{(k)}, \cdots, \bW_J^{(k)} }$ denotes the set of all blocks except block $j$, at iteration $k$.
While many choices are possible for the update order of the blocks, we adopt the cyclic update rule, where $k$ satisfies $j = k \mod J $, to simplify the presentation. Thus, BCD results in the the following subproblem, at iteration $k$
\begin{align}
(Q)
\begin{cases}
\underset{ \pmb{W}_j \in \calW_j }{\argmin}~f_j(  \bW_j ; \bW_{-j}^{(k)} ) , ~ k = 1,2,\cdots
\end{cases}
\end{align}
Recall that $ f_j(  \bW_j ; \bW_{-j}^{(k)} )$ denotes the regularized loss, when `looking' only at $\bW_j$.  Note further that $f_j(  \bW_j ; \bW_{-j}^{(k)} )$ is a composition of non-linear activation functions, followed by $\ell$.
Since the activation and the loss function are assumed to be differentiable, then $f_j(  \bW_j ; \bW_{-j}^{(k)} )$ is differentiable - though not necessarily convex. In general, $(Q)$ is not a convex problem, implying that standard BCD methods are not applicable (convergence cannot be guaranteed). Instead, we leverage recent methods such as, Inexact Flexible Parallel Descent~\citep{Facchinei_PABDO_TSP_14} and Block-Successive Upperbound Minimization (BSUM)~\citep{Razaviyayn_BCD_12} where $(Q)$ is solved via a series of carefully designed convex upperbounds on $f_j(  \bW_j ; \bW_{-j}^{(k)} )$.

\begin{figure}
\centering
  \includegraphics[height=4cm, width=8cm ]{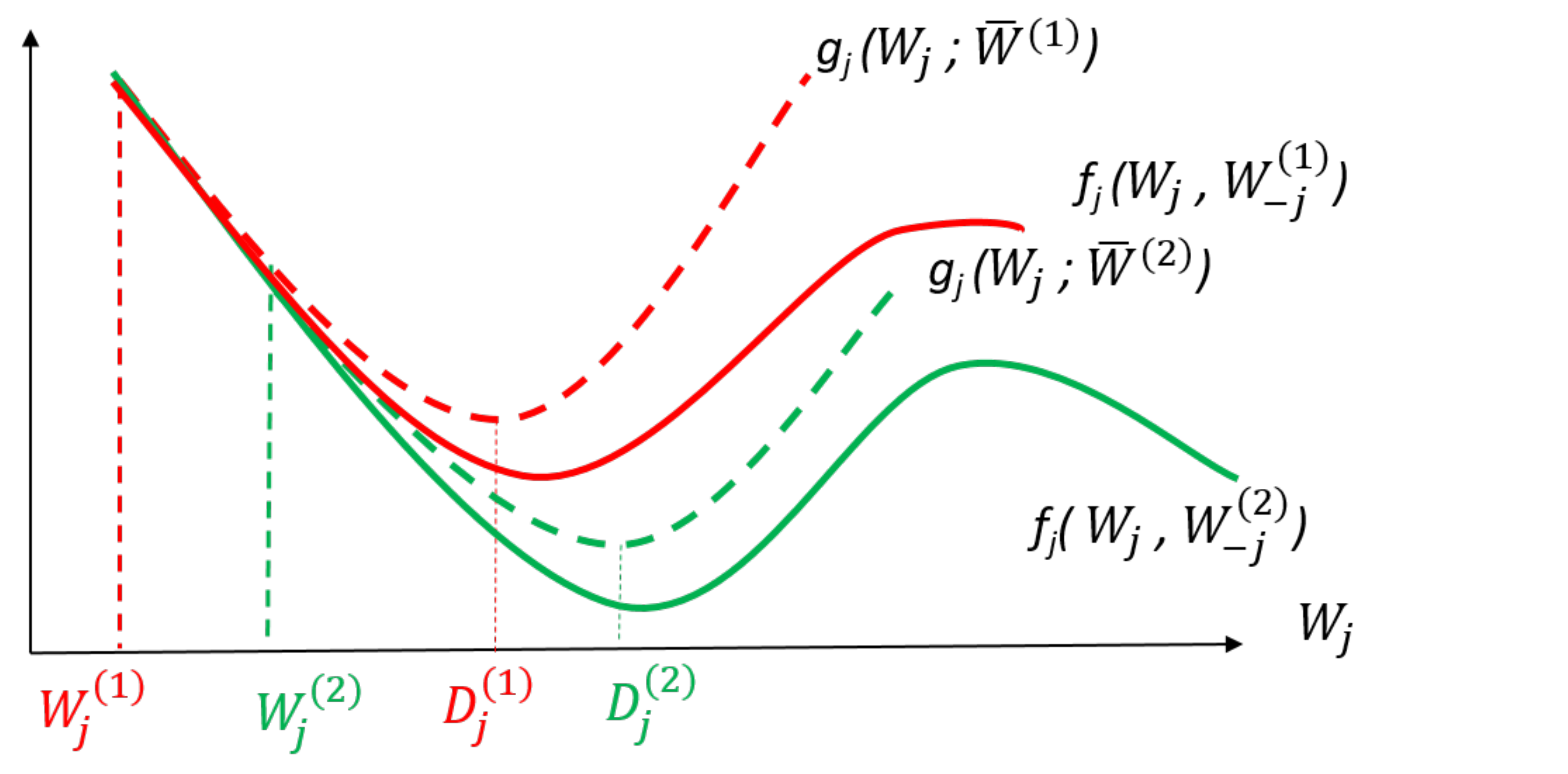}
  \caption{Illustration of Proposed Approach for $\bW_j$ (2 iterations, $k=1$ in red, $k=2$ in green). Given $\bW_j^{(1)}$, $g_j(\bW_j ; \bbW^{(k)}) $ is an upperbound for $ f_j(  \bW_j ; \bW_{-j}^{(1)} )$. The descent direction $\bD_j^{(1)}$ is the minimizer of $g_j(\bW_j ; \bbW^{(1)}) $. $\bW_j^{(2)}$ is a convex combination of $\bW_j^{(1)}$ and $\bD_j^{(1)}$. $\bW_j^{(2)}$ is the starting point of second iteration. }
  \label{fig:BSUM}
  \vspace{-.5cm}
\end{figure}

Let $g_j(\bW_j ; \bbW^{(k)} ) : \IR^{d_j} \times \IR^{d_{j-1}} \rightarrow \IR $ denote a given convex upperbound of $f_j$, at point $\bbW^{(k)}$, where $\bbW^{(k)} = \lrb{ \bW_1^{(k)} , \cdots, \bW_J^{(k)} }$. Though many choices are available for the upperbound, they must satisfy some properties (Appendix~\ref{app:UB_prop}).
Following the method proposed by~\citet{Facchinei_PABDO_TSP_14}, a \emph{surrogate} problem for $(Q)$ is posed, where $f_j$ is upperbounded by its convex surrogate, $g_j$, to find the\emph{ descent direction} $\bD_j^{(k)}$ for block $\bW_j^{(k)}$: 
\begin{align} \label{opt:upd_Zk}
\tilde{(Q)}: ~ \bD_j^{(k)} \triangleq
\begin{cases}
\underset{ \pmb{W}_j  \in \calW_j }{\argmin}~g_j(  \bW_j ; \bW_{-j}^{(k)} )  , ~ k = 1,2,\cdots
\end{cases}
\end{align}
The strong convexity of $g_j(\bW_j ; \bbW^{(k)} )$ and $\calW_j$ (by construction) imply that $\tilde{(Q)}$ is strongly convex, and its solution $\bD_j^{(k)}$ is unique.
Then, $\bW_j$ is updated as a \emph{convex combination} of its current value and the descent direction, as follows:
\begin{align} \label{eq:upd_Wk}
\bW_j^{(k+1)} = (1-\alpha_j^{(k)})\bW_j^{(k)} +  \alpha_j^{(k)} \bD_j^{(k)} , ~
\end{align}
where $\alpha_j^{(k)}$ is the \emph{stepsize} satisfying~\citep{Facchinei_PABDO_TSP_14}:
\begin{align} \label{eq:stepcond}
0 \leq \alpha_j^{(k)} < 1 , ~ \lim_{k \rightarrow \infty} \alpha_j^{(k)} = 0, ~\sum_k \alpha_j^{(k)} = \infty  , ~ \sum_k \alpha_j^{(k)^2} < \infty, \forall j \in \calJ ~.
\end{align}
Other choices for $\alpha_j^{(k)}$ include \emph{small constant}~\citep{Razaviyayn_PSCA_14} or using \emph{Armijo's rule}~\citep{Facchinei_PABDO_TSP_14}. The approach is illustrated in Figure~\ref{fig:BSUM}.

\subsection{Algorithm Description}
Once an upperbound is selected, during each iteration the algorithm updates the descent direction, the stepsize, and the corresponding weights,  until convergence; see Algorithm~\ref{alg:BSUM}.
Next, we show the algorithms' convergence to a stationary point of $(P)$.
\begin{algorithm}
  \caption{Training with Alternating Minimization  } \label{alg:BSUM}
  \begin{algorithmic}
    \For{$k=1,2,...,$}
    \State Set $j=k \mod J$ (for the cyclic update) 
    \State Find descent direction $\bD_j^{(k)}$ by solving \eqref{opt:upd_Zk}
    \State Update stepsize $\alpha_j^{(k)}$ (satisfying~\eqref{eq:stepcond}) 
    \State Update $\bW_j^{(k)}$ by solving~\eqref{eq:upd_Wk}
    \EndFor
  \end{algorithmic}
\end{algorithm}
\vspace{-.5cm}

\begin{lemma}[Convergence of Algorithm~\ref{alg:BSUM}]. \label{lem:conv} \rm
Let $\lrb{ \bbW^{(k)} \triangleq (\bW_1^{(k)} , \cdots, \bW_J^{(k)} )}_k$ denote the sequence of iterates generated by the updates in~\eqref{opt:upd_Zk} and~\eqref{eq:upd_Wk}, where $\alpha_j^{(k)}$ satisfies the conditions in~\eqref{eq:stepcond}. Then, every limit point of $\lrb{\bbW^{(k)}}$ is a stationary point of $(P)$.  
\end{lemma}
While the proof is based on a stepsize that satisfies~\eqref{eq:stepcond}, convergence still holds in the case of a constant step size~\citep{Razaviyayn_PSCA_14}[Theorem 1].  $\lrb{f (\bbW^{(k)}) }_k $, \emph{is not necessarily}  monotonically decreasing: only asymptotic converge of $\lrb{\bbW^{(k)}} $ to a stationary point of $(P)$, as $ k \rightarrow \infty$, is guaranteed~\citep{Razaviyayn_PSCA_14,Facchinei_PABDO_TSP_14}. 
Note that, Lemma~\ref{lem:conv} only relies on minimal assumptions (Assumptions~\ref{ass:loss},\ref{ass:activation}). Thus, convergence holds for almost all loss functions (except the $0/1$ loss), regularization functions, and smooth activation functions (though the result can be extended for piecewise linear functions). Consequently, with a proper choice of upperbounds, we can apply our general algorithm to most DNN learning tasks.
Additionally, when $f$ is strongly convex in each of its  blocks (see following discussion in Section~\ref{sec:fj_concave}), the iterations of Algorithm~\ref{alg:BSUM} \emph{monotonically converge} to the stationary point, following standard BCD converge~\citep{Tseng_convBCD_01}.



\section{Choice of Upperbounds} \label{sec:UBchoice}
In this section, we explore several choices for the upperbounds $g_j$. Each of these choices yields a different cost function for $\tilde{(Q)}$, and consequently, different updates for $\bD_j^{(k)}$ and $\bW_j^{(k+1)}$.
We focus on the first-order proximal upperbound (next) and proximal upperbound (Section~\ref{sec:prox_UB}) in the main text. Derivations on the second-order and linear upperbounds are included in Appendix~\ref{app:SOPA}, and Appendix~\ref{app:linUB}, respectively.
For simplicity, we assume that $\tilde{(Q)} $ is unconstrained.\footnote{When $\tilde{(Q)}$ is constrained with a closed convex set, $\calW_j$, it is a strongly convex problem that can be solved with any standard convex optimization method (KKT conditions, proximal point). The most suitable method should be selected depending on the specific problem instance.}

\subsection{First-order Proximal Upperbound}
We start with a first-order proximal upperbound,
\begin{align} \label{eq:g_j_FOPA}
g_j(\bW_j ; \bbW^{(k)} ) &= f(\bbW^{(k)}) + \tr \left[ \nabla_{\bW_j} f(\bbW^{(k)})^T (\bW_j - \bW_j^{(k)} ) \right] +  \frac{\gamma_j}{ 2 } \Vert \bW_j - \bW_j^{(k)} \Vert_F^2  ~,
\end{align}
where $\nabla_{\bW_j} f(\bbW^{(k)}) = \nabla_{\bW_j} f(\bbW) \vert_{ \bbW = \bbW^{(k)} } $ is the gradient of $f$ with respect to $\bW_j$ at $\bW_j^{(k)}$, and $\gamma_j > 0$ is the Lipschitz constant for $g_j(\bW_j ; \bbW^{(k)} )$.
Notice that, the upperbound $g_j$ is evaluated at point $\bbW^{(k)}$, and controlled by the design parameter $\gamma_j$.
From~\eqref{eq:g_j_FOPA}, we can see that $\tilde{(Q)}$ is a strongly convex problem, which can be solved by finding the stationary point of the Lagrangian, resulting in the following updates:
\begin{align} \label{opt:upd_reg_FOA}
 \bD_j^{(k)} &= \bW_j^{(k)} - \gamma_j^{-1} \nabla_{\bW_j} f(\bbW^{(k)}) , \nonumber \\
 \bW_j^{(k+1)} &=   \bW_j^{(k)} -  \alpha_j^{(k)} \gamma_j^{-1} \nabla_{\bW_j} f(\bbW^{(k)}) .
\end{align}
The gradient may be computed using the chain rule of calculus; see Appendix~\ref{sec:app:grad_deriv}.
Note that the update in~\eqref{opt:upd_reg_FOA} is applicable to all loss functions satisfying Assumption~\ref{ass:loss}. In Appendix~\ref{app:SOPA} we extended the derivations of this section to a second-order upperbound. 

\subsubsection{Convergence of First- and Second-order methods}
Here we can see how the updates resulting from first and second order upperbounds, i.e.,~\eqref{opt:upd_reg_FOA} and~\eqref{opt:upd_reg_SOA} (Appendix~\ref{app:SOPA}) are generalizations of well-known methods, e.g., \emph{back propagation (BP)} and its variants, \emph{gradient descent}, and the \emph{Newton-Raphson method}.

\paragraph{Special Case 1 (BackPropagation):}
Let us first consider the case where we set $\gamma_j = 1$ in~\eqref{opt:upd_reg_FOA}. Then, the proposed algorithm reduces to the \emph{BP} method, and $ \alpha_j^{(k)}$  becomes the \emph{learning rate}.
Recall that different choices of the learning rate have lead to many popular variants of BP, such as constant, decaying, and  ADAGRAD.
A common drawback of these approaches for choosing the learning rate is the lack of convergence guarantees in most DNN training problems: Therefore it is usually selected via cross validation~\citep{Zeiler_ADADELTA_12}.
With the proposed algorithm, however, the convergence in Lemma~\ref{lem:conv} holds for any choice of stepsize satisfying~\eqref{eq:stepcond}. 

\begin{corollary}[Convergence of BP]. \rm \label{cor:convFoM}
For any arbitrary smooth loss $f$ (Assumptions~\ref{ass:loss},~\ref{ass:activation}) the updates of first-order methods in~\eqref{opt:upd_reg_FOA} converge to a stationary point of~$(P)$. This includes BP and many of its variants, e.g., BP with constant but small $ \alpha_j^{(k)} $, inverse-root $ \alpha_j^{(k)} = c/\sqrt{k} ~(c>0)$, Arimijo's rule, etc., as they satisfy~\eqref{eq:stepcond}.
The proof  follows directly from Lemma~\ref{lem:conv}.
\end{corollary}

We underline that convergence of BP for a smooth loss function, for the conditions in~\eqref{eq:stepcond}, was shown by~\citet{Mangasarian_BPConv_93}[Theorem 2.1]. However, the result of Corollary~\ref{cor:convFoM} also extends to the Newton-Raphson method derived in Appendix~\ref{app:SOPA}.


\paragraph{Special Case 2 (Gradient Descent and Newton-Raphson Method):}
Another notable special case of our proposed algorithm is obtained by considering a unit stepsize, i.e., setting $\alpha_j^{(k)} = 1$~in~\eqref{opt:upd_reg_FOA}. Then, the update for  $\bW_j$ in \eqref{opt:upd_reg_FOA} reduces to,
\begin{align} \label{eq:SD}
 \bW_j^{(k+1)} = \bW_j^{(k)} - \gamma_j^{-1} \nabla_{\bW_j} f(\bbW^{(k)})  ,
\end{align}
which is the \emph{gradient descent}, with stepsize $\frac{1 }{\gamma_j}$.
Similarly, when $\alpha_j^{(k)} = 1$, the update resulting from the second-order upperbound in~\eqref{opt:upd_reg_SOA} reduces to the Levenberg-Marquardt variant of the  Newton-Raphson method~\citep{Yu_LM_11},
\begin{align} \label{eq:Netwon}
 \bW_j^{(k+1)} = \bW_j^{(k)} - \left[\nabla_{\bW_j}^2 f(\bbW^{(k)})  + {\gamma_j} \bI\right]^{-1} \nabla_{\bW_j} f(\bbW^{(k)})  .
\end{align}
Evidently, it is quite desirable to have a convergence proof for these training methods.
To our best knowledge, such results are not available in the literature, in the general setting considered here. Note that while setting $\alpha_j^{(k)} = 1$ violates the stepsize conditions in~\eqref{eq:stepcond}, converge of \eqref{eq:SD} and~\eqref{eq:Netwon} follow from the BSUM framework (in the result below).
\vspace{-.2cm}
\begin{lemma}[Convergence of Gradient/Newton Methods].  \label{lem:conv_SD_Newt} \rm
The updates generated by the gradient method in \eqref{eq:SD}, or Newton's method in~\eqref{eq:Netwon}, converges to a stationary point of $(P)$.
\end{lemma}
\vspace{-.2cm}
\textbf{Discussions:} In that sense, our proposed approach \emph{complements} known training methods (e.g., gradient descent, BP and its variants, Newton), by providing conditions for convergence guarantees.
We underline that first- and second-order methods (i.e., BP, gradient descent, Newton’s method) can be interpreted as special cases of the BSUM method (when using first- and second-order upperbounds only), as
 convex approximations of a (non-convex) loss function around each block.

\textbf{Practical Issues:}
The updates resulting from the first-order proximal upperbound require knowledge of the gradient (resp. Hessian)  of $f$ with respect to $\bW_j$: They are  similar to BP (resp. LMBFS) methods for training DNNs, in terms of the computational complexity needed to run the algorithms. Moreover, the stochastic training variant (Section~\ref{sec:stoch_train}) may be used when the training set is too large.
Finally, the fact that the algorithm is based on BCD/BSUM methods enables  parallel processing, to make algorithm distributed~\citep{Lorenzo_DistML_SCA_16}.

\subsubsection{From Batch to Stochastic Training}  \label{sec:stoch_train}
In contemporary machine learning tasks, its quite common to deal with large datasets for which the computation of the gradient over all samples is challenging. In such scenarios, we can replace the gradient with its unbiased noisy estimate, and run the following update:
\begin{align}
\hat{\bW}_j^{(k+1)} &= \hat{\bW}_j^{(k)} - \alpha_j^{(k)} \gamma_j^{-1} \widehat{\nabla_{\bW_j}}f(\bbW^{(k)}) ,~
\end{align}
where the \emph{stochastic (or mini-batch) gradient}, $\widehat{\nabla_{\bW_j}}f$, is defined in~\eqref{eq:SGD_BP}.
Following the results from~\citep{Facchinei_PABDO_TSP_14}, if the error is bounded, $\Vert \hat{\bW}_j^{(k)} -  \bW_j^{(k)}  \Vert_F^2 \leq \epsilon_j^{(k)} $, and $\epsilon_j^{(k)}$ decays asymptotically to zero, then the convergence of Algorithm~\ref{alg:BSUM} still holds. A complete convergence proof of this stochastic training variation is left for future work.

\subsection{Proximal Upperbound:} \label{sec:prox_UB}
When $f_j (\bW_j ; \bW_{-j}^{(k)} ) $ is convex (discussed next), the proximal operator is yet another possible upperbound. Unlike the first-order upperbound,
the descent direction is the solution to a proximal-point operator with respect to $\gamma_j^{-1} f_j$, around $\bW_j^{(k)}$, as follows:
\begin{align} \label{eq:proxUB}
\bD_j^{(k)} = \textup{prox}_{{\gamma_j^{-1}} f_j } ( \bW_j^{(k)} ) = \underset{ \pmb{W}_{j} \in \calW_j  }{\argmin} ~f_j (\bW_j ; \bW_{-j}^{(k)} )  +  \frac{\gamma_j}{ 2 } \Vert \bW_j - \bW_j^{(k)} \Vert_F^2 ~.
\end{align}
Notice that the update in~\eqref{eq:proxUB} are a general form of the proximal algorithm by \citet{Frerix_ProxProp_ICLR18}, developed for the regression problem.

\subsubsection{Convexity of $f$ in Each Block } \label{sec:fj_concave}
The convexity of $f$ in each of its blocks will impact the algorithm and analytical results. When $f_j(  \bW_j ; \bW_{-j}^{(k)} )$ is strongly convex (as shown below), then simple BCD methods may be employed instead of upperbounds. 
In the following, we find conditions on the choice of $\ell$ and $\bSig_j$, that result in $f_j(  \bW_j ; \bW_{-j}^{(k)} )$ being convex. From~\eqref{eq:f_jdef}, note that $f_j$ can be abstracted as:
\begin{align} \label{eq:fj_composition}
f_j(  \bW_j ; \bW_{-j}^{(k)} ) &= \ell \left[  \bSig_J( \bW_j^{(k)} \cdots \bSig_j( \bW_j  \cdots \bSig_1(\bW_1^{(k)} ))) \right] + r_j(\bW_j) \nonumber \\
&= \ell \left[  \bSig_J(  \cdots \bSig_j( \bW_j   )) \right] + r_j(\bW_j) .
\end{align}
The second equality follows from noting that composition with an affine function does not alter the convexity/concavity of $f_j$~\citep{boyd_convex_2004}[Chap 3.2.4], implying that the fixed blocks, $\bW_{-j}^{(k)}$, may be dropped without altering this analysis: Indeed, $f_j$ is a composition of activation functions $\bSig_j, \cdots, \bSig_J $, followed by $\ell$. We next find two alternate sufficient conditions, that guarantee the strong convexity of $f$ in each of its blocks.
\vspace{-.3cm}
\begin{proposition}[Convexity of $f_j$]. \label{prop:f_j_convex} \rm
$f_j(  \bW_j ; \bW_{-j}^{(k)} )$ is strongly convex in $\bW_j$ with constant $\lambda_j$, if \newline
C1) $\lrb{\bSig_j}_{j=1}^J $ is convex and non-decreasing, and $\ell$ is convex and non-decreasing, or  \newline
C2) $\lrb{\bSig_j}_{j=1}^J $ is concave and non-decreasing, and is $\ell$ convex and non-increasing.
\end{proposition}
\vspace{-.3cm}
While the conditions of C1) are satisfied in many cases (most activation function are convex and non-decreasing), those of C2) are stricter.
This result implies that one can engineer convex formulations (via specific choices of $\ell$ and $r_j$), where standard BCD methods can be used to solve each subproblem \emph{exactly} (since $(Q)$ becomes strongly convex).
We elaborate on these combinations in Appendix~\ref{app:comb_loss_activ}.
Similarly, choices of $\ell$ and $\bSig_j$ that lead to $f_j$ being concave in each block are discussed in Appendix~\ref{app:concave_fj}.


\section{Applications} \label{sec:application}
\subsection{Training DNNs} \label{sec:Regress}
\paragraph{Regression Tasks:}
The problem of regression consists of optimizing the non-linear network, to `best fit' a given training set. For a ridge regression, the loss function and regularization are the $l_2$-norm ~\citep{Bishop_PatternRec_06}:
\begin{align} \label{eq:RegDNN}
f(\bbW) = \frac{1}{N} \Vert \bY - \bH(\bX) \Vert_F^2 + \sum_j \lambda_j \Vert \bW_j \Vert_F^2 , ~ \lambda_j > 0 .
\end{align}
 Moreover, the feasible set for $\bW_j$ simply becomes $\calW_j = \IR^{d_j \times d_{j-1} } $. Sparse  regularization (LASSO) is discussed in Appendix~\ref{app:noncovex_reg}. Essentially, we can run Algorithm~\ref{alg:BSUM} for loss functions such as~\eqref{eq:RegDNN}, with a guaranteed convergence.

\paragraph{Classification Tasks:} \label{sec:class}
The proposed approach is also applicable to DNN classification tasks, where the
regularized cross-entropy loss, in~\eqref{eq:CrossEnt}, is prevalent.
Now, we can run Algorithm 1 on~~\eqref{eq:CrossEnt}. Our approach is equally applicable to other classification  loss functions, e.g., the regularized squared hinge loss~\eqref{eq:SqHinge}, and the logistic loss~\eqref{eq:logistic} (detailed in Appendix \ref{sec:app:grad_deriv}).


\subsection{Training Various DNN Architectures}
\paragraph{Deep Convolutional Neural Networks:}
 While the methods for regression and classification (Section~\ref{sec:Regress}) were in the context of classical DNNs, they may be modified to train deep convolutional neural networks. In this setting, the weights for each layer are a Toeplitz matrix, representing the circular convolution operation, where the rows of $ \bW_j $ are taken from the set of $d_j \times d_{j-1} $ Toeplitz matrices, $\calT_j $.
Moreover, it is known that $\calT_j$ is a closed affine subspace in $\bW_j$~\citep{Eberle_ToeplitzProj_03}. Thus, $\tilde{(Q)}$ reduces to $\argmin_{\bW_j \in \calT_j}~g_j(  \bW_j ; \bW_{-j}^{(k)} )$,
where $g_j(  \bW_j ; \bW_{-j}^{(k)} )$ can be chosen as any of the upperbounds in Section~\ref{sec:UBchoice}, and Appendix~\ref{app:SOPA}. Furthermore, $\tilde{(Q)}$ is strongly convex, since $\calT_j$ is a affine.

\paragraph{Deep Linear Networks: }
Deep linear networks are recent attempts at making DNNs (mathematically) tractable, by using identity activation functions; see Section~\ref{sec:Intro}. For such system, the model consists of a cascade of linear operators, $\bH(\bZ_0) \triangleq ~ (\bW_J \cdots ~ \bW_1) \bZ_0 $. The proposed method is indeed applicable to training these networks. More specifically, BCD methods are applicable  in this case, since $(Q)$ is a strongly convex problem, and can be solved in closed-form.

\section{Numerical Results} \label{sec:numres}
\textbf{Simulation setting:} We present simple numerical results to \emph{validate} the various convergence claims, thereby focusing on \emph{training} rather than test performance.
 Recall that known methods such as BP with decaying learning rate ($\alpha_j^{(k)}= c/\sqrt{k}$) are instances of the proposed method. We also explore other choices of the stepsize~, i.e., $\alpha^{(k+1)} = \alpha_j^{(k)} ( 1 - t \alpha_j^{(k)} ), 0 < t < 1$~\citep{Facchinei_PABDO_TSP_14}, and $\alpha_j^{(k)}= c/2^{k}$, all of which satisfy the stepsize conditions in~\eqref{eq:stepcond}. We benchmark against other approaches such as, BP with constant learning rate (BP-CLR), and ADAGRAD~\citep{Zeiler_ADADELTA_12}.
We use the BodyFat regression dataset ($N=252$) for a small DNN (to avoid overfitting) with $J=4$, $d_0 = 13, d_1 = d_2 = d_3 = 10$, and $ d_4 = 1 $. Moreover, the $l_2$ loss in~\eqref{eq:RegDNN} and logistic activation function are used.  

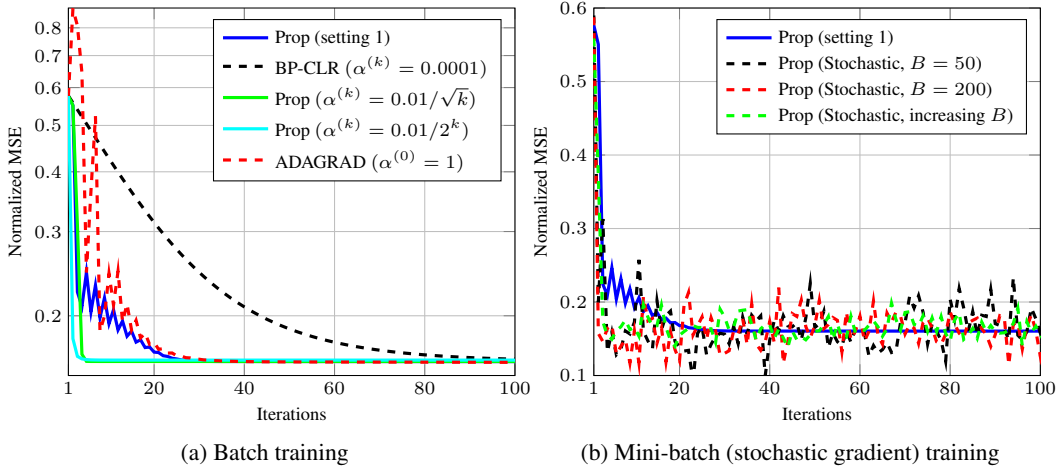
\begin{figure*}[t!]
    \centering
    \begin{subfigure}[t]{0.5\textwidth}
        \centering
 {\scriptsize\input{./figs/BF_training}}
        \caption{Batch training}
        \label{fig:trn_BF_batch}
    \end{subfigure}%
    \begin{subfigure}[t]{0.5\textwidth}
        \centering
    {\scriptsize\input{./figs/BF_train_batch_vs_stock}}
        \caption{Mini-batch (stochastic gradient) training }
        \label{fig:trn_BF_stoch}
    \end{subfigure}
    \caption{Normalized MSE performance for the proposed schemes and benchmark. ``Prop'' stands for the proposed approach. ``setting 1'' is $\alpha^{(k+1)} = \alpha^{(k)}( 1 - 0.99\alpha^{k}),\alpha^{(0)}= 1$. In \subref{fig:trn_BF_stoch}, the increasing $B$ curve corresponds to $B = \min(k,N)$.}
    \vspace{-.5cm}
\end{figure*}

Figure~\ref{fig:trn_BF_batch} shows the normalized training mean-squared error (MSE), for the various training methods.
Moreover, all algorithms converge to a stationary point (as shown in Corollary~\ref{cor:convFoM}, except for ADAGRAD). Note that the oscillations are still consistent with Lemma~\ref{lem:conv}, as only asymptotic convergence is ensured.
We observe that all three variants of the proposed method outperform BP-CLR and ADAGRAG (in terms of convergence rate).
An interesting by-product of our framework is the possibility of selecting the best learning algorithm, in terms of convergence rate.
Using the same optimization framework, we can implement various algorithms and pick the best, all of which having convergence guarantees.

Recall from Proposition~\ref{prop:f_j_convex} and its related discussions, that we can achieve monotonic convergence when $f$ is strongly convex in each block. The latter is feasible by adopting among others, an exponential loss~(given by~\eqref{eq:exploss}), and a SoftPlus activation. Indeed, condition C1) in Proposition~\ref{prop:f_j_convex} holds in this case, and consequently, monotonic convergence can be shown (using standard BCD convergence), as seen from Figure~\ref{fig:trn_BF_convblock} in Appendix~\ref{app:add_numres}).

Figure~\ref{fig:trn_BF_stoch} additionally shows the training MSE of the stochastic variant, in Section~\ref{sec:stoch_train}, each with a different batch size $B$. We observe that the performance improves as $B$ increases from $50$ to $200$, as expected. In the last variant, $B$ is increased with the number of iterations (until $B=N$): Thus the variance in the gradient estimate is gradually reduced, allowing for convergence. Note that the results are similar to the convergence behavior of stochastic gradient methods.

\section{Conclusion}
In this work, we proposed a framework for training general neural networks and showed its convergence using the BSUM method, for a wide class of DNNs architectures and learning tasks.
We showed that updates resulting from first- (resp. second-) order proximal upperbound were in fact generalizations of well-known first-order (resp. Newton) methods.
Thus, the proposed method allows us to analyze and show convergence of these methods, for training DNNs. While these connections arise from our investigation of just first- and second-order upperbounds, we wish to explore in the future other upperbounds, which may lead to different algorithms, and further applications (e.g., robustness in learning).

\vskip 0.2in
\newpage
\bibliography{ref_hadi_merged}
\bibliographystyle{unsrtnat}

\cleardoublepage

\appendix
\centerline{\LARGE{\textbf{Supplementary Material for}}}\vspace{2mm}
\centerline{\LARGE{\textbf{A Unified Framework for Training Neural Networks}}}\vspace{5mm}

\setcounter{page}{1}
\setcounter{equation}{0}
\setcounter{figure}{0}
\renewcommand{\theequation}{A.\arabic{equation}}
\renewcommand\thefigure{A.\arabic{figure}}

\section{Additional Results}
\subsection{Conditions for convex $f_j$} \label{app:comb_loss_activ}
Choices of loss and activation functions for which Condition C1) holds (in Proposition~\ref{prop:f_j_convex}).
Activation functions such as, differentiable variants of the ReLU (leaky, parametric, exponential, scaled exponential, inverse square-root), softplus, Bent identity, and soft exponential. Moreover, the exponential loss in~\eqref{eq:exploss} is convex and monotonically increasing.

The conditions for C2) holds (in Proposition~\ref{prop:f_j_convex}) are stricter, as they are satisfied by log-type activation functions (a subset of soft exponential), and most loss functions such as, smoothed or squared hinge, logistic, and the cross-entropy.

\subsection{Conditions for concave $f_j$} \label{app:concave_fj}
Note that one may also identify choices of loss and activation function that result in $f$ being concave in each of its blocks: this is case of class for the class of exponentially concave functions, which include the logistic loss in classification~\citep{Mahdavi_ExpConcLoss_14}, studied by~\citep{Mahdavi_ExpConcLoss_14}, \citep{Pal_ExpConc_InfoGeo_16}. We refer the reader to the survey~\citep{Mehta_ExpConcLearning_16}.

\begin{proposition} \label{rmk:fj_concave} \rm
 Consider the case  $\lrb{\bSig_j}_{j=1}^J $ are convex and non-decreasing, $\ell$ is concave a non-increasing, and no regularization is added to $\ell$. Then, $f_j$ is concave in each block (the proof follows a similar argument as Proposition~\ref{prop:f_j_convex}). In such cases, the gradient can be used as upperbound. Moreover, overfitting can be prevented by including the regularization as a (convex) constraint, $\calW_j = \lrb{  r_j(\bW_j) \leq \rho_j } $.
\end{proposition}

\subsection{Properties of Upperbounds} \label{app:UB_prop}
Below we formalize the analytical properties of the upperbound~s\citep{Facchinei_PABDO_TSP_14}:
\begin{itemize}
\item[P1:] $g_j(\bW_j ; \bbW^{(k)} )$ is continuously differentiable in $\bW_j$ and has a Lipschitz continuous gradient with constant $ \gamma_j$.
\item[P2:] $ \nabla_{\bW_j} ~g_j(\bW_j ; \bbW^{(k)} )  = \nabla_{\bW_j} ~f(\bW_1, \cdots, \bW_J) $
\item[P3:] $g_j(\bW_j ; \bbW^{(k)} )$ has a continuous gradient on $\calW_j$
\item[P4:] $g_j(\bW_j ; \bbW^{(k)} )$ is strongly convex on $\calW_j$, with Lipschitz constant $ \gamma_j $
\end{itemize}

\subsection{Second-order Upperbound:} \label{app:SOPA}
Note that tighter approximations can be done using the Hessian.
Naturally, we need to make the added assumptions that $\ell$ and $r_j$ are twice differentiable (in this part  only). The second-order proximal upperbound is defined as,
\begin{align} \label{eq:g_j_SOPA}
g_j(\bW_j ; \bbW^{(k)} ) &= f(\bbW^{(k)}) + \tr [ \nabla_{\bW_j} f(\bbW^{(k)})^T (\bW_j - \bW_j^{(k)} ) ] + \frac{\gamma_j}{2} \Vert \bW_j - \bW_j^{(k)}  \Vert_F^2 \nonumber \\
&~~ + \frac{1}{2} \tr [ (\bW_j - \bW_j^{(k)} )^T \nabla_{\bW_j}^2 f(\bbW^{(k)})  (\bW_j - \bW_j^{(k)} )   ]  .
\end{align}
where $\nabla_{\bW_j}^2 f(\bbW^{(k)})$ is the Hessian of $f$ with respect to $\bW_j$, at $\bbW^{(k)}$.
We can drop the constant invariant terms in the last equation,
\begin{align*}
 g_j(\bW_j ; \bbW^{(k)} ) &= \tr [ \nabla_{\bW_j} f(\bbW^{(k)})^T \bW_j] \\ 
& ~~ + \frac{1}{2} \tr \left[ (\bW_j - \bW_j^{(k)} )^T ( \nabla_{\bW_j}^2 f(\bbW^{(k)}) + {\gamma_j} \bI ) (\bW_j - \bW_j^{(k)} )\right]
\end{align*}

With this upperbound $\tilde{(Q)}$ is strongly convex. Using Lagrangian techniques, we derive its solution to obtain the descent direction, and updates: 
\begin{align} \label{opt:upd_reg_SOA}
 \bD_j^{(k)} &= \bW_j^{(k)} - [\nabla_{\bW_j}^2 f(\bbW^{(k)}) + {\gamma_j} \bI]^{-1} \nabla_{\bW_j} f(\bbW^{(k)})  \nonumber \\
\bW_j^{(k+1)} &=   \bW_j^{(k)} -  \alpha_j^{(k)} [\nabla_{\bW_j}^2 f(\bbW^{(k)}) + {\gamma_j} \bI]^{-1}  \nabla_{\bW_j} f(\bbW^{(k)})
\end{align}
We recall that the convergence in Corollary~\ref{cor:convFoM} still holds for the update in~\eqref{opt:upd_reg_SOA}.

\subsection{Linear Upperbound:} \label{app:linUB}
Note that when $f_j$ is concave in each of its blocks (Appendix~\ref{app:concave_fj}), the upperbound in~\eqref{eq:g_j_FOPA} reduces to a linear one, i.e.,
$g_j(\bW_j ; \bbW^{(k)} ) = f(\bbW^{(k)}) + \tr [ \nabla_{\bW_j} f(\bbW^{(k)})^T (\bW_j - \bW_j^{(k)} ) ] ,$
yields a descent direction $\bD_j^{(k)} = \nabla_{\bW_j} f(\bbW^{(k)})  $, and the following update:
\begin{align} \label{eq:upd_linUB}
 \bW_j^{(k+1)} &=  (1-\alpha_j^{(k)}) \bW_j^{(k)} - \alpha_j^{(k)}  \nabla_{\bW_j} f(\bbW^{(k)})
\end{align}
Following the discussion in Appendix~\ref{app:concave_fj}, we highlight the wide class of exponentially concave functions as direct applications of the update in~\eqref{eq:upd_linUB}, for future investigations.
Moreover, its convergence follows from Lemma~\ref{lem:conv}.

\subsection{Extension to Non-smooth Regularization} \label{app:noncovex_reg}
Note that the proposed approach can be extended to cases where regularization is non-smooth but convex, i.e., $r_j (\bW_j) = \lambda_j \Vert \bW_j \Vert_1 $, where $\Vert \bA \Vert_1 = \sum_{(n,m)} |\bA_{n,m}|  $ . For illustration purposes, we restrict the derivations to the first-order approximations in~\eqref{eq:g_j_FOPA} of the smooth loss . However, the discussion below is applicable to all other upperbounds presented in the work. In the non-smooth regularization case~\eqref{eq:g_j_FOPA} becomes,
\begin{align}
g_j(\bW_j ; \bW_{-j}^{(k)} ) &= \ell(\bbW^{(k)}) + \tr [ \nabla_{\bW_j} \ell(\bbW^{(k)})^T (\bW_j - \bW_j^{(k)} ) ] +  \frac{\gamma_j}{ 2 } \Vert \bW_j - \bW_j^{(k)} \Vert_F^2
\end{align}
The descent direction is determined as,
\begin{align}
\bD_j^{(k+1)} &\triangleq \underset{\bW_j}{\argmin}~g_j(\bW_j ; \bW_{-j}^{(k)} ) + r_j (\bW_j)= \underset{\bW_j}{\argmin}~g_j(\bW_j ; \bW_{-j}^{(k)} ) + \lambda_j \Vert \bW_j \Vert_1 \nonumber \\
&=  \textup{prox}_{ \gamma_j^{-1} r_j } [\bW_j^{(k)} - \gamma_j^{-1} \nabla_{\bW_j}~g_j( \bbW^{(k)} )]
\end{align}
where $\textup{prox}$ is the proximal operator with respect to $ \gamma_j^{-1} r_j $ , given by the soft thresholding operation~\citep{Parikh_ProxAlg_14}:
\begin{align*}
[ \textup{prox}_{ \frac{1 }{ \gamma_j} r_j  } (\bA)]_{n,m} =
 \begin{cases}
[\bA]_{n,m} - \frac{\lambda_j }{ \gamma_j} &, ~  [\bA]_{n,m} \geq \frac{\lambda_j }{ \gamma_j} \\
0 ~&, - \frac{\lambda_j }{ \gamma_j} \leq [\bA]_{n,m} \leq \frac{\lambda_j }{ \gamma_j} \\
[\bA]_{n,m} + \frac{\lambda_j }{ \gamma_j} ~&, [\bA]_{n,m} \leq \frac{\lambda_j }{ \gamma_j}
 \end{cases} .
\end{align*}

\subsection{Additional Numerical Results} \label{app:add_numres}
\begin{figure}[h]
\centering
    {\scriptsize\input{./figs/BF_training_convexblck1}}
   \caption{Normalized MSE performance for the proposed schemes and benchmark. ``Prop'' stands for the proposed approach. ``setting 2'' is $\alpha^{(k+1)} = \alpha^{(k)}( 1 - 0.9\alpha^{k}),\alpha^{(0)}= 0.001$.}
   \label{fig:trn_BF_convblock}
\end{figure}
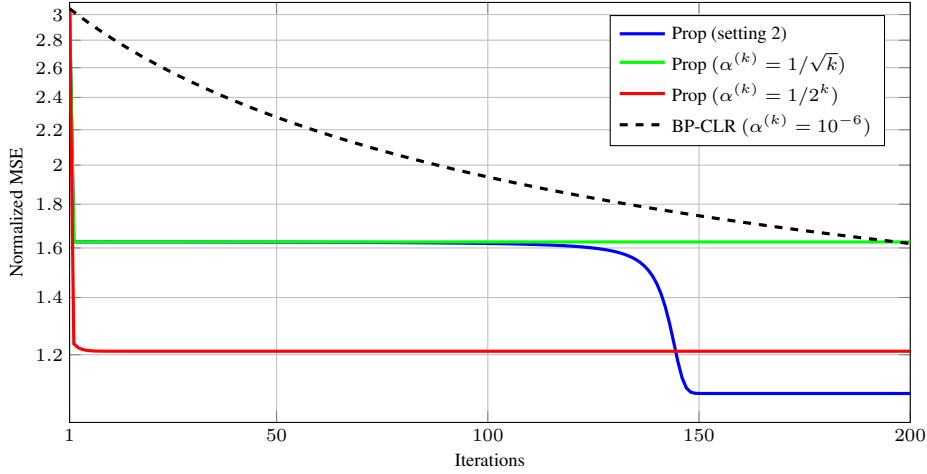

Using the same parametrization of~Section~\ref{sec:numres}, we change exponential loss~(given by~\eqref{eq:exploss}), and a SoftPlus activation. The resulting loss is convex in each block of coordinates. Convergence of the various algorithms is shown in  Figure~\ref{fig:trn_BF_convblock}.  

\section{Additional Derivations}
\subsection{First-order Derivations} \label{sec:app:grad_deriv}
We provide (brief) derivations of the gradient of $f$ for arbitrary $\ell$ and $r_j$.
We included them in matrix form to avoid cluttering the notation with indexes. We denote by $\nabla_{\bW_j} f \in \IR^{d_j \times d_{j-1} } $ the gradient of the regularized loss $f$ with respect to $\bW_j$,   $\nabla_{\bW_j} r_j $ is the gradient of $r_j(\bW_j)$. and let $\bU_j = \bW_j \bZ_{j-1} $. After simple manipulations, we write
\begin{align} \label{eq:grad_f}
\nabla_{\bW_j} f = \bDel_j \bZ_{j-1}^T + \nabla_{\bW_j} r_j ~, \forall j \in {J}
\end{align}
where $\bDel_j \in \IR^{ d_j \times N } $ is defined recursively as,
\begin{align} \label{eq:Delta_j}
 \bDel_j \in \IR^{d_j \times N}, \bDel_j=
\begin{cases}
   \nabla_{\bH} \ell \circ \bSig_J^{'}(\bU_j)  ~ &, \textup{ for } j = J \\
  (\bW_{j+1}^T \bDel_{j+1}) \circ \bSig_j^{'}(\bU_j) ~ &, \textup{ for } j < J
\end{cases} ,
\end{align}

 Moreover, $\nabla_{\bH} \ell \in \IR^{ d_J \times N } $ is the gradient of the loss $\ell$, and $\circ$ is the element-by-element Hadamard product, and $\bSig_j^{'}(\bU_j) \in \IR^{d_j \times N} $ is gradient of the element-by-element operation $\bSig_j(\bU_j)$: 
\begin{align}
\bSig_j^{'} (\bU_j) \triangleq
\begin{bmatrix}
\frac{ \partial \sigma_{1,1}  }{ \partial u_{1,1} } & \cdots & \frac{ \partial \sigma_{1,N}  }{ \partial u_{1,N} } \\
  & \vdots &  \\
\frac{ \partial \sigma_{d_j ,1}  }{ \partial u_{d_j,1} } & \cdots & \frac{ \partial \sigma_{d_j,N}  }{ \partial u_{d_j,N} }
\end{bmatrix}
\end{align}

The above derivations are obtained by reformulating the standard backpropagation equations, e.g.,~\citet{Haykin_NNC_1998}[Chap. 4],
\begin{align} \label{eq:gradBP}
\nabla_{\bW_j} f = \frac{1}{N} \sum_{n=1}^N \bdel_j^n \bz_{j-1}^{n^T} + \nabla_{\bW_j} r_j ~, \forall j \in {J}
\end{align}
where the summand is the gradient of $\ell$ for sample $(\bx_n, \by_n)$, $\bz_j^n \in \IR^{d_j} $ the output of layer $j$ corresponding to input $ \bx_n$. Moreover, $\bdel_j^n \in \IR^{d_j}$, $ \bdel_j^n = (\bW_{j+1}^T \bdel_{j+1}^n) \circ \bsig_j^{n'}, \textup{ for } j < J $, and $ \bdel_J^n = \nabla_{\bH} \ell \circ \bsig_J^{n'}  $.
Then, \eqref{eq:grad_f}, \eqref{eq:Delta_j} are obtained by letting $ \bDel_j = [\bdel_j^1 , ..., \bdel_j^N] $  and $ \bZ_j = [\bZ_j^1 , ..., \bz_j^N] $. Similarly to the gradient over all training samples,~\eqref{eq:gradBP}, the stochastic gradient over a batch of samples $\calB \subseteq \lrb{N} $ is defined as:
\begin{align} \label{eq:SGD_BP}
\widehat{\nabla_{\bW_j}} f = \frac{1}{\vert \calB \vert} \sum_{n \in \calB}\bdel_j^n \bz_{j-1}^{n^T} + \nabla_{\bW_j} r_j ~, \forall j \in {J}
\end{align}

One can then apply these results directly to the many problems investigated here, by simply plugging in the corresponding values for $\nabla_{\bH} \ell$ and $\nabla_{\bW_j} r_j$, in~\eqref{eq:grad_f} and~\eqref{eq:Delta_j}.

\paragraph{$l_2$ loss/regularization:}
In the case of training DNNs and DCNNs for regression, i.e.,~\eqref{eq:RegDNN}, we have
\begin{align}
 \nabla_{\bH} \ell = \frac{2}{N}(\bH(\bX) - \bY )  \textup{ and } \nabla_{\bW_j} r_j = 2 \lambda_j \bW_j
\end{align}

\paragraph{Exponential Loss:}
\begin{align} \label{eq:exploss}
f= c \exp^{ \frac{1}{c}~ \frac{1}{N} \Vert \bY - \bH(\bX) \Vert_F^2} + \sum_{j} \lambda_j \Vert \bW_j \Vert_F^2 ,
\end{align}
for some non-negative constant $c$. Moreover, the gradient is given by
\begin{align}
 \nabla_{\bH} \ell = 2 (\bH(\bX) - \bY ) \exp^{ \frac{1}{c}~ \Vert \bY - \bH(\bX) \Vert_F^2}  \textup{ and } \nabla_{\bW_j} r_j = 2 \lambda_j \bW_j
\end{align}

\paragraph{Cross-entropy loss:}
\begin{align} \label{eq:CrossEnt}
f = - \frac{1}{N}\left[ \bY \circ \log \left( \bH(\bX) \right) + ( \pmb{\Omega} - \bY) \circ \log \left( \pmb{\Omega}- \bH(\bX) \right) \right] + \sum_j \lambda_j \Vert \bW_j \Vert_F^2
\end{align}
where $\bY \in \IB^{d_J \times N} $ for classification, $\pmb{\Omega} \in \IR^{d_J \times N} $ is matrix of all ones,  and $\log(\bA)$ is an element-by-element operator on $\bA$.
In the case of~\eqref{eq:CrossEnt}, we have
\begin{align}
 \nabla_{\bH} \ell &=  \frac{1}{N} \left[ - \bY \circ \frac{1}{ \bH(\bX) } + (\pmb{\Omega} - \bY) \circ \left(\frac{1}{ \pmb{\Omega} - \bH(\bX) } \right) \circ \bH(\bX)\right] \\
  \textup{ and } &\nabla_{\bW_j} r_j = 2 \lambda_j \bW_j
\end{align}
where $ \frac{1}{\bA} $ is the element-by-element inverse of $\bA$.

\paragraph{Squared-hinge loss:}
\begin{align} \label{eq:SqHinge}
f = \frac{1}{2cN} ( \pmb{\Omega} - \bY \circ \bH(\bX)  )_+^2  + \sum_j \lambda_j \Vert \bW_j \Vert_F^2  ,
\end{align}
where $c$ is a non-negative constant and $(\bA)_+ $ is the $\max(0,x)$ applied to each element of $\bA$.

\paragraph{Logistic Loss:}
\begin{align} \label{eq:logistic}
f = \frac{1}{N} \sum_{n=1}^N \log(1 + e^{ - \by_n^T \bH(\bx_n) } ) + \sum_j \lambda_j \Vert \bW_j \Vert_F^2
\end{align}
 where $\log(\bA)$ are applied element-wise. 

\section{Proofs}
\subsection{Proof of Proposition~\ref{prop:f_j_convex}} \label{app:prooffjconvex}
Though written in matrix form, the result relies on scalar composition (since the activation functions are element-by-element operators).
Recall that $f_j(\bW_j ; \bW_{-j}^{(k)} ) = \ell \left[  \bSig_J(  \cdots \bSig_j( \bW_j   )) \right] + r_j(\bW_j) $.
Thus, the convexity of $f$ in each of it blocks is equivalent to showing that $f_j(\bW_j ; \bW_{-j}^{(k)} ) $ is convex (the main idea behind the proof). Note that the convexity of $f_j(\bW_j ; \bW_{-j}^{(k)} ) $ depends on the composition of   $\ell \left[  \bSig_J(  \cdots \bSig_j( \bW_j   )) \right]$ and  $r_j(\bW_j)$ (from the discussion around Proposition~\ref{prop:f_j_convex}).

\textbf{Condition C1):} Note that  $  \bSig_J( \cdots \bSig_j( \bW_j) ) $  is a composition of convex and non-decreasing functions. Thus, $  \bSig_J( \cdots \bSig_j( \bW_j) ) $ is convex in $\bW_j$~(Property~\ref{property:compfunc} below). Moreover, since $\ell$ is convex and monotonically increasing, $ \ell( \bSig_J( \cdots \bSig_j( \bW_j) ) ) $  is convex \citep{boyd_convex_2004}[Chap 3.2.4]. Consequently, $ f_j = \ell( \bSig_J( \cdots \bSig_j( \bW_j) ) )  + r_j(\bW_j)$ is strongly convex, since $r_j$ is strongly convex (Assumption~\ref{ass:loss}).

\textbf{Condition C2):} If $  \bSig_J( \cdots \bSig_j( \bW_j) ) $  is a composition of concave non-decreasing functions, then $  \bSig_J( \cdots \bSig_j( \bW_j) ) $ is concave in $\bW_j$~(Property~\ref{property:compfunc} below). Moreover, if $\ell$ is convex and non-decreasing, then $\ell( \bSig_J( \cdots \bSig_j( \bW_j) ) )$ is convex in $\bW_j$ \citep{boyd_convex_2004}[Chap 3.2.4]. Thus the convex loss $\ell$, combined with and the strongly convex $r_j$, implies that $f_j$ is strongly convex in $\bW_j$.

\begin{property}: \label{property:compfunc} \rm
Let $\phi_1(x), ..., \phi_n(x) $ be a collection of convex (resp. concave) non-decreasing functions. Then, $\phi_n(... \phi_1(x) ) $ is convex (resp. concave). The proof follows directly~\citep{boyd_convex_2004}[Chap 3.2.4].
\end{property}

\subsection{Proof of Lemma~\ref{lem:conv} }
The proof is a special case of \citep{Facchinei_PABDO_TSP_14}[Theorem 1], where only exact solutions are found in each block, and the non-smooth part is set to zero.

\subsection{Proof of Lemma~\ref{lem:conv_SD_Newt} }
The convergence results from~Lemma~\ref{lem:conv} require that $\alpha_j^{(k)} < 1 $, which is incompatible with the above updates. However, one can use convergence results of the BSUM method from~\citep{Razaviyayn_BCD_12}, when a simple cyclic update rule is used, and a single block is updated~\citep{Razaviyayn_PSCA_14}.

\end{document}

%% file: Packages.tex
\usepackage[utf8]{inputenc} 
\usepackage[T1]{fontenc}    
\usepackage{hyperref}       
\usepackage{url}            
\usepackage{booktabs}       
\usepackage{amsfonts,amsmath, amstext, amssymb, mathrsfs, graphicx, color, url}
\usepackage{nicefrac}       
\usepackage{microtype}      
\usepackage{algpseudocode}
\usepackage{algorithm}
\usepackage{epstopdf}
\usepackage{graphicx,wrapfig}
\usepackage{titlesec}
\usepackage[labelformat=simple]{subcaption}

\let\subparagraph\relax
\titlespacing*{\section}
{0pt}{2pt}{2pt}
\titlespacing*{\subsection}
{0pt}{2pt}{2pt}

\makeatletter
\g@addto@macro\normalsize{%
  \setlength\abovedisplayskip{2pt}
  \setlength\belowdisplayskip{1pt}
  \setlength\abovedisplayshortskip{2pt}
  \setlength\belowdisplayshortskip{1pt}
}

\usepackage{pgf}
\usepackage{pgfplots}
\usepackage{tikz}
\usepackage{grffile}
\pgfplotsset{compat=newest}
\usetikzlibrary{plotmarks}
\usetikzlibrary{quotes,angles,automata,arrows,positioning,calc,3d,matrix,decorations.markings}

%% file: Macros.tex
\def\IB{{\mathbb B}}

\def\IR{{\mathbb R}}

\def\calB{{\mathcal B}}

\def\calJ{{\mathcal J}}

\def\calT{{\mathcal T}}

\def\calW{{\mathcal W}}

\def\bA{\mbox {\boldmath $A$}}

\def\bD{\mbox {\boldmath $D$}}

\def\bH{\mbox {\boldmath $H$}}
\def\bI{\mbox {\boldmath $I$}}

\def\bU{\mbox {\boldmath $U$}}

\def\bW{\mbox {\boldmath $W$}}
\def\bX{\mbox {\boldmath $X$}}
\def\bY{\mbox {\boldmath $Y$}}
\def\bZ{\mbox {\boldmath $Z$}}

\def\bbW{\overline{{\bW}}}

\def\bx{\mbox {\boldmath $x$}}
\def\by{\mbox {\boldmath $y$}}
\def\bz{\mbox {\boldmath $z$}}

\def\bSig{\mbox{\boldmath $\Sigma$}}

\def\bDel{\mbox{\boldmath $\Delta$}}

\def\bsig{\mbox{\boldmath $\sigma$}}
\def\bdel{\mbox{\boldmath $\delta$}}

\newcommand{\upperRomannumeral}[1]{\uppercase\expandafter{\romannumeral#1}}

\newcommand{\argmin}{ \textup{argmin} }

\def\tr{{\textup{tr}}}

\newcommand{\lrb}[1]{ \lbrace #1 \rbrace }

\def\({\left(}
\def\){\right)}

\newtheorem{lemma}{Lemma}

\newtheorem{proposition}{Proposition}

\newtheorem{corollary}{Corollary}

\newtheorem{property}[lemma]{Property}

\newtheorem{assumption}{Assumption}

%% file: figs/BF_training.tex
\definecolor{mycolor1}{rgb}{0.00000,1.00000,1.00000}%
\begin{tikzpicture}

\begin{axis}[%
width=0.85\columnwidth,
height=0.7\columnwidth,
at={(1.037in,0.517in)},
scale only axis,
xmin=1,
xmax=100,
xlabel={Iterations},
ymode=log,
ymin=0.15,
ymax=0.88,
yminorticks=true,
xtick={1,20,40,60,80,100},
ytick={0.2,0.3,0.4,0.5,0.6,0.7,0.8},
ylabel={Normalized MSE},
ylabel near ticks,
xmajorgrids,
ymajorgrids,
log ticks with fixed point,
axis background/.style={fill=white},
legend columns=1,
legend style={legend pos=north east,legend cell align=left,align=left,draw=black}
]
\addplot [color=blue, line width=1.2pt]
  table[row sep=crcr]{%
1	0.576582433691381\\
2	0.550564732104159\\
3	0.22419889030256\\
4	0.209094194358271\\
5	0.246431718643538\\
6	0.203838132071617\\
7	0.231613858497913\\
8	0.198242918104625\\
9	0.21803230351955\\
10	0.192498977691283\\
11	0.205822213048274\\
12	0.186776448783835\\
13	0.1950576747057\\
14	0.181262611680581\\
15	0.185812078352353\\
16	0.176154330740969\\
17	0.178142977274781\\
18	0.171644571345731\\
19	0.172070037891538\\
20	0.167897264444067\\
21	0.167545230104265\\
22	0.165011796807147\\
23	0.164426259682356\\
24	0.162989186571348\\
25	0.162472589679403\\
26	0.161721857681252\\
27	0.161379132348964\\
28	0.161023428288172\\
29	0.160839578609784\\
30	0.16068872244867\\
31	0.160606629663208\\
32	0.160549705168571\\
33	0.160518408358629\\
34	0.160499028204812\\
35	0.160488290715308\\
36	0.160481874265028\\
37	0.160478032408148\\
38	0.160475469591093\\
39	0.160473577353175\\
40	0.160472006601338\\
41	0.160470591855088\\
42	0.16046925137319\\
43	0.160467948092551\\
44	0.160466665573199\\
45	0.160465396761775\\
46	0.160464138763426\\
47	0.160462890348225\\
48	0.16046165099476\\
49	0.160460420428302\\
50	0.160459198480366\\
51	0.160457985013538\\
52	0.160456779906647\\
53	0.16045558304317\\
54	0.160454394310735\\
55	0.160453213599208\\
56	0.160452040800931\\
57	0.160450875810327\\
58	0.16044971852396\\
59	0.160448568840408\\
60	0.160447426660238\\
61	0.160446291885946\\
62	0.160445164421908\\
63	0.160444044174332\\
64	0.160442931051213\\
65	0.160441824962287\\
66	0.160440725818989\\
67	0.160439633534406\\
68	0.160438548023243\\
69	0.160437469201777\\
70	0.160436396987825\\
71	0.160435331300701\\
72	0.160434272061181\\
73	0.160433219191471\\
74	0.160432172615172\\
75	0.160431132257242\\
76	0.160430098043973\\
77	0.16042906990295\\
78	0.160428047763029\\
79	0.160427031554303\\
80	0.160426021208075\\
81	0.160425016656829\\
82	0.160424017834207\\
83	0.160423024674978\\
84	0.160422037115014\\
85	0.160421055091268\\
86	0.160420078541746\\
87	0.160419107405488\\
88	0.160418141622542\\
89	0.160417181133941\\
90	0.160416225881685\\
91	0.160415275808719\\
92	0.160414330858912\\
93	0.160413390977036\\
94	0.160412456108748\\
95	0.160411526200574\\
96	0.160410601199885\\
97	0.160409681054884\\
98	0.160408765714586\\
99	0.160407855128802\\
100	0.160406949248124\\
101	0.160406048023906\\
102	0.160405151408251\\
103	0.160404259353993\\
104	0.160403371814686\\
105	0.160402488744587\\
106	0.160401610098642\\
107	0.160400735832471\\
108	0.16039986590236\\
109	0.16039900026524\\
110	0.160398138878682\\
111	0.160397281700878\\
112	0.160396428690633\\
113	0.160395579807352\\
114	0.160394735011029\\
115	0.160393894262234\\
116	0.160393057522103\\
117	0.160392224752328\\
118	0.160391395915146\\
119	0.160390570973327\\
120	0.160389749890168\\
121	0.160388932629479\\
122	0.160388119155575\\
123	0.160387309433268\\
124	0.160386503427856\\
125	0.160385701105115\\
126	0.160384902431289\\
127	0.160384107373086\\
128	0.160383315897662\\
129	0.16038252797262\\
130	0.160381743565998\\
131	0.160380962646263\\
132	0.160380185182302\\
133	0.160379411143416\\
134	0.160378640499313\\
135	0.160377873220097\\
136	0.160377109276268\\
137	0.160376348638707\\
138	0.160375591278678\\
139	0.160374837167814\\
140	0.160374086278113\\
141	0.160373338581936\\
142	0.160372594051994\\
143	0.160371852661347\\
144	0.160371114383395\\
145	0.160370379191875\\
146	0.160369647060855\\
147	0.160368917964725\\
148	0.160368191878196\\
149	0.160367468776292\\
150	0.160366748634348\\
151	0.160366031428\\
152	0.160365317133184\\
153	0.160364605726131\\
154	0.160363897183358\\
155	0.16036319148167\\
156	0.160362488598151\\
157	0.160361788510158\\
158	0.160361091195323\\
159	0.160360396631542\\
160	0.160359704796975\\
161	0.160359015670039\\
162	0.160358329229407\\
163	0.160357645454001\\
164	0.160356964322991\\
165	0.160356285815788\\
166	0.160355609912043\\
167	0.160354936591642\\
168	0.160354265834704\\
169	0.160353597621573\\
170	0.160352931932821\\
171	0.160352268749238\\
172	0.160351608051836\\
173	0.160350949821838\\
174	0.160350294040679\\
175	0.160349640690005\\
176	0.160348989751664\\
177	0.160348341207707\\
178	0.160347695040386\\
179	0.160347051232145\\
180	0.160346409765627\\
181	0.160345770623661\\
182	0.160345133789265\\
183	0.160344499245643\\
184	0.160343866976181\\
185	0.160343236964444\\
186	0.160342609194174\\
187	0.160341983649289\\
188	0.160341360313878\\
189	0.1603407391722\\
190	0.160340120208681\\
191	0.160339503407913\\
192	0.160338888754649\\
193	0.160338276233804\\
194	0.160337665830449\\
195	0.160337057529814\\
196	0.16033645131728\\
197	0.16033584717838\\
198	0.160335245098798\\
199	0.160334645064364\\
200	0.160334047061054\\
};
\addlegendentry{Prop (setting 1)}

\addplot [color=black, dashed, line width=1.2pt]
  table[row sep=crcr]{%
1	0.576582433691381\\
2	0.557687366475856\\
3	0.539306908664382\\
4	0.521462620246653\\
5	0.50417242352357\\
6	0.487450600658898\\
7	0.471307847168858\\
8	0.455751376215356\\
9	0.4407850674867\\
10	0.426409653712898\\
11	0.412622937467702\\
12	0.399420030833294\\
13	0.386793610707993\\
14	0.374734182974082\\
15	0.363230349358268\\
16	0.352269071556831\\
17	0.341835928009512\\
18	0.331915359544433\\
19	0.322490900941909\\
20	0.313545396247117\\
21	0.305061196377669\\
22	0.297020338207312\\
23	0.289404704853239\\
24	0.282196167349583\\
25	0.275376708255742\\
26	0.268928528030768\\
27	0.262834135212012\\
28	0.257076421576546\\
29	0.251638723547253\\
30	0.246504871141555\\
31	0.241659225758541\\
32	0.237086708068293\\
33	0.232772817212802\\
34	0.228703642457581\\
35	0.224865868352287\\
36	0.22124677437186\\
37	0.217834229920492\\
38	0.214616685491907\\
39	0.211583160693075\\
40	0.208723229756201\\
41	0.206027005086492\\
42	0.203485119321706\\
43	0.201088706313853\\
44	0.198829381384052\\
45	0.196699221148086\\
46	0.194690743162622\\
47	0.192796885599922\\
48	0.191010987121867\\
49	0.189326767091815\\
50	0.187738306234816\\
51	0.186240027832564\\
52	0.184826679518822\\
53	0.183493315723468\\
54	0.18223528079851\\
55	0.181048192846922\\
56	0.179927928264873\\
57	0.178870606999343\\
58	0.177872578516217\\
59	0.176930408468301\\
60	0.176040866048322\\
61	0.175200912008469\\
62	0.174407687325455\\
63	0.17365850248813\\
64	0.172950827383369\\
65	0.172282281755116\\
66	0.17165062621101\\
67	0.171053753750936\\
68	0.170489681791988\\
69	0.1699565446647\\
70	0.169452586555955\\
71	0.168976154874644\\
72	0.168525694016909\\
73	0.168099739508655\\
74	0.167696912503895\\
75	0.167315914618411\\
76	0.166955523079154\\
77	0.166614586170725\\
78	0.166292018961231\\
79	0.165986799290708\\
80	0.165697964006197\\
81	0.165424605428442\\
82	0.165165868035993\\
83	0.164920945353344\\
84	0.164689077030457\\
85	0.164469546101825\\
86	0.164261676413873\\
87	0.164064830210221\\
88	0.163878405864937\\
89	0.163701835754511\\
90	0.163534584259881\\
91	0.163376145890352\\
92	0.163226043521769\\
93	0.163083826741777\\
94	0.162949070295479\\
95	0.162821372625176\\
96	0.162700354498323\\
97	0.162585657718184\\
98	0.16247694391201\\
99	0.162373893391919\\
100	0.162276204083943\\
101	0.162183590521014\\
102	0.162095782895905\\
103	0.162012526170433\\
104	0.161933579237432\\
105	0.16185871413224\\
106	0.161787715290669\\
107	0.161720378850579\\
108	0.16165651199441\\
109	0.161595932330146\\
110	0.161538467308384\\
111	0.161483953673303\\
112	0.161432236945484\\
113	0.161383170934645\\
114	0.161336617280492\\
115	0.161292445019994\\
116	0.161250530179493\\
117	0.161210755390164\\
118	0.161173009525428\\
119	0.161137187359018\\
120	0.161103189242467\\
121	0.161070920800866\\
122	0.161040292645816\\
123	0.161011220104567\\
124	0.16098362296438\\
125	0.160957425231237\\
126	0.160932554902046\\
127	0.160908943749568\\
128	0.160886527119317\\
129	0.160865243737745\\
130	0.160845035531063\\
131	0.160825847454075\\
132	0.160807627328459\\
133	0.160790325689955\\
134	0.160773895643942\\
135	0.160758292728931\\
136	0.160743474787525\\
137	0.160729401844426\\
138	0.160716035991078\\
139	0.160703341276585\\
140	0.160691283604544\\
141	0.160679830635462\\
142	0.160668951694448\\
143	0.160658617683878\\
144	0.160648801000762\\
145	0.160639475458551\\
146	0.160630616213134\\
147	0.160622199692799\\
148	0.160614203531932\\
149	0.160606606508255\\
150	0.160599388483403\\
151	0.160592530346663\\
152	0.160586013961692\\
153	0.160579822116066\\
154	0.160573938473488\\
155	0.160568347528529\\
156	0.160563034563746\\
157	0.160557985609067\\
158	0.160553187403304\\
159	0.160548627357688\\
160	0.160544293521316\\
161	0.160540174548407\\
162	0.160536259667269\\
163	0.160532538650886\\
164	0.160529001789037\\
165	0.160525639861866\\
166	0.160522444114833\\
167	0.160519406234955\\
168	0.160516518328288\\
169	0.160513772898573\\
170	0.160511162826985\\
171	0.160508681352929\\
172	0.160506322055832\\
173	0.160504078837869\\
174	0.160501945907579\\
175	0.160499917764328\\
176	0.160497989183567\\
177	0.160496155202849\\
178	0.160494411108563\\
179	0.160492752423346\\
180	0.160491174894143\\
181	0.160489674480877\\
182	0.160488247345696\\
183	0.160486889842769\\
184	0.160485598508606\\
185	0.160484370052871\\
186	0.160483201349657\\
187	0.160482089429213\\
188	0.160481031470088\\
189	0.160480024791673\\
190	0.160479066847124\\
191	0.160478155216642\\
192	0.160477287601097\\
193	0.160476461815973\\
194	0.160475675785616\\
195	0.160474927537781\\
196	0.160474215198448\\
197	0.160473536986901\\
198	0.16047289121106\\
199	0.160472276263043\\
200	0.160471690614955\\
};
\addlegendentry{BP-CLR $(\alpha^{(k)}= 0.0001)$}

\addplot [color=green, line width=1.2pt]
  table[row sep=crcr]{%
1	0.576582433691381\\
2	0.550564732104159\\
3	0.308608338772717\\
4	0.166113478989776\\
5	0.160585966345729\\
6	0.16050490230848\\
7	0.160503713456134\\
8	0.160502958376258\\
9	0.160502252488828\\
10	0.160501587054486\\
11	0.160500955838942\\
12	0.160500354062499\\
13	0.160499777963712\\
14	0.160499224519424\\
15	0.160498691256625\\
16	0.160498176121772\\
17	0.160497677387499\\
18	0.160497193584405\\
19	0.160496723450153\\
20	0.1604962658908\\
21	0.160495819950932\\
22	0.16049538479029\\
23	0.160494959665231\\
24	0.160494543913858\\
25	0.160494136943966\\
26	0.160493738223188\\
27	0.160493347270854\\
28	0.160492963651229\\
29	0.160492586967842\\
30	0.160492216858705\\
31	0.160491852992254\\
32	0.16049149506389\\
33	0.160491142792998\\
34	0.160490795920393\\
35	0.160490454206087\\
36	0.160490117427363\\
37	0.160489785377081\\
38	0.160489457862194\\
39	0.16048913470244\\
40	0.160488815729189\\
41	0.160488500784421\\
42	0.160488189719807\\
43	0.160487882395905\\
44	0.160487578681426\\
45	0.160487278452587\\
46	0.160486981592516\\
47	0.160486687990735\\
48	0.160486397542668\\
49	0.16048611014922\\
50	0.160485825716376\\
51	0.160485544154845\\
52	0.160485265379736\\
53	0.160484989310261\\
54	0.160484715869459\\
55	0.16048444498395\\
56	0.160484176583704\\
57	0.160483910601834\\
58	0.160483646974397\\
59	0.160483385640221\\
60	0.160483126540734\\
61	0.16048286961982\\
62	0.160482614823667\\
63	0.160482362100645\\
64	0.160482111401182\\
65	0.160481862677647\\
66	0.160481615884251\\
67	0.160481370976945\\
68	0.160481127913329\\
69	0.160480886652571\\
70	0.16048064715532\\
71	0.160480409383639\\
72	0.160480173300928\\
73	0.160479938871866\\
74	0.160479706062344\\
75	0.160479474839409\\
76	0.160479245171212\\
77	0.160479017026953\\
78	0.160478790376837\\
79	0.160478565192027\\
80	0.160478341444601\\
81	0.160478119107513\\
82	0.160477898154552\\
83	0.160477678560313\\
84	0.160477460300155\\
85	0.160477243350174\\
86	0.160477027687171\\
87	0.160476813288624\\
88	0.160476600132659\\
89	0.160476388198027\\
90	0.160476177464075\\
91	0.160475967910727\\
92	0.160475759518461\\
93	0.160475552268285\\
94	0.160475346141721\\
95	0.160475141120782\\
96	0.160474937187956\\
97	0.160474734326189\\
98	0.160474532518868\\
99	0.160474331749803\\
100	0.160474132003216\\
101	0.160473933263724\\
102	0.160473735516325\\
103	0.160473538746386\\
104	0.160473342939631\\
105	0.160473148082126\\
106	0.160472954160272\\
107	0.160472761160789\\
108	0.160472569070711\\
109	0.16047237787737\\
110	0.160472187568391\\
111	0.16047199813168\\
112	0.160471809555417\\
113	0.160471621828046\\
114	0.160471434938266\\
115	0.160471248875027\\
116	0.160471063627518\\
117	0.160470879185162\\
118	0.160470695537608\\
119	0.160470512674725\\
120	0.160470330586596\\
121	0.160470149263511\\
122	0.160469968695959\\
123	0.160469788874628\\
124	0.160469609790393\\
125	0.160469431434313\\
126	0.160469253797628\\
127	0.160469076871751\\
128	0.160468900648265\\
129	0.160468725118918\\
130	0.160468550275617\\
131	0.160468376110426\\
132	0.160468202615561\\
133	0.160468029783385\\
134	0.160467857606404\\
135	0.160467686077266\\
136	0.160467515188755\\
137	0.160467344933787\\
138	0.160467175305408\\
139	0.16046700629679\\
140	0.160466837901228\\
141	0.160466670112139\\
142	0.160466502923053\\
143	0.160466336327618\\
144	0.16046617031959\\
145	0.160466004892836\\
146	0.160465840041328\\
147	0.160465675759142\\
148	0.160465512040454\\
149	0.160465348879539\\
150	0.160465186270769\\
151	0.160465024208608\\
152	0.160464862687613\\
153	0.160464701702432\\
154	0.160464541247799\\
155	0.160464381318533\\
156	0.160464221909538\\
157	0.160464063015801\\
158	0.160463904632385\\
159	0.160463746754436\\
160	0.160463589377173\\
161	0.160463432495891\\
162	0.160463276105959\\
163	0.160463120202815\\
164	0.160462964781971\\
165	0.160462809839004\\
166	0.16046265536956\\
167	0.160462501369351\\
168	0.160462347834151\\
169	0.160462194759799\\
170	0.160462042142195\\
171	0.1604618899773\\
172	0.160461738261134\\
173	0.160461586989774\\
174	0.160461436159355\\
175	0.160461285766066\\
176	0.160461135806153\\
177	0.160460986275912\\
178	0.160460837171696\\
179	0.160460688489904\\
180	0.16046054022699\\
181	0.160460392379454\\
182	0.160460244943847\\
183	0.160460097916766\\
184	0.160459951294854\\
185	0.160459805074801\\
186	0.160459659253342\\
187	0.160459513827254\\
188	0.160459368793359\\
189	0.160459224148521\\
190	0.160459079889644\\
191	0.160458936013676\\
192	0.160458792517602\\
193	0.160458649398447\\
194	0.160458506653276\\
195	0.16045836427919\\
196	0.160458222273328\\
197	0.160458080632867\\
198	0.160457939355017\\
199	0.160457798437025\\
200	0.160457657876173\\
};
\addlegendentry{Prop $(\alpha^{(k)}=0.01/\sqrt{k})$}

\addplot [color=mycolor1, line width=1.2pt]
  table[row sep=crcr]{%
1	0.576582433691381\\
2	0.179132934286151\\
3	0.164497453970717\\
4	0.162502619125525\\
5	0.161947808207809\\
6	0.161736890749677\\
7	0.161644517353229\\
8	0.161601244024018\\
9	0.161580295318479\\
10	0.161569988156462\\
11	0.161564875788136\\
12	0.161562329834926\\
13	0.161561059407088\\
14	0.161560424829243\\
15	0.1615601076992\\
16	0.16155994917388\\
17	0.161559869921144\\
18	0.161559830297257\\
19	0.161559810485933\\
20	0.161559800580426\\
21	0.161559795627711\\
22	0.161559793151364\\
23	0.161559791913192\\
24	0.161559791294107\\
25	0.161559790984565\\
26	0.161559790829794\\
27	0.161559790752408\\
28	0.161559790713715\\
29	0.161559790694369\\
30	0.161559790684696\\
31	0.161559790679859\\
32	0.161559790677441\\
33	0.161559790676232\\
34	0.161559790675627\\
35	0.161559790675325\\
36	0.161559790675174\\
37	0.161559790675098\\
38	0.16155979067506\\
39	0.161559790675042\\
40	0.161559790675032\\
41	0.161559790675027\\
42	0.161559790675025\\
43	0.161559790675024\\
44	0.161559790675023\\
45	0.161559790675023\\
46	0.161559790675023\\
47	0.161559790675023\\
48	0.161559790675023\\
49	0.161559790675023\\
50	0.161559790675023\\
51	0.161559790675023\\
52	0.161559790675023\\
53	0.161559790675023\\
54	0.161559790675023\\
55	0.161559790675023\\
56	0.161559790675023\\
57	0.161559790675023\\
58	0.161559790675023\\
59	0.161559790675023\\
60	0.161559790675023\\
61	0.161559790675023\\
62	0.161559790675023\\
63	0.161559790675023\\
64	0.161559790675023\\
65	0.161559790675023\\
66	0.161559790675023\\
67	0.161559790675023\\
68	0.161559790675023\\
69	0.161559790675023\\
70	0.161559790675023\\
71	0.161559790675023\\
72	0.161559790675023\\
73	0.161559790675023\\
74	0.161559790675023\\
75	0.161559790675023\\
76	0.161559790675023\\
77	0.161559790675023\\
78	0.161559790675023\\
79	0.161559790675023\\
80	0.161559790675023\\
81	0.161559790675023\\
82	0.161559790675023\\
83	0.161559790675023\\
84	0.161559790675023\\
85	0.161559790675023\\
86	0.161559790675023\\
87	0.161559790675023\\
88	0.161559790675023\\
89	0.161559790675023\\
90	0.161559790675023\\
91	0.161559790675023\\
92	0.161559790675023\\
93	0.161559790675023\\
94	0.161559790675023\\
95	0.161559790675023\\
96	0.161559790675023\\
97	0.161559790675023\\
98	0.161559790675023\\
99	0.161559790675023\\
100	0.161559790675023\\
101	0.161559790675023\\
102	0.161559790675023\\
103	0.161559790675023\\
104	0.161559790675023\\
105	0.161559790675023\\
106	0.161559790675023\\
107	0.161559790675023\\
108	0.161559790675023\\
109	0.161559790675023\\
110	0.161559790675023\\
111	0.161559790675023\\
112	0.161559790675023\\
113	0.161559790675023\\
114	0.161559790675023\\
115	0.161559790675023\\
116	0.161559790675023\\
117	0.161559790675023\\
118	0.161559790675023\\
119	0.161559790675023\\
120	0.161559790675023\\
121	0.161559790675023\\
122	0.161559790675023\\
123	0.161559790675023\\
124	0.161559790675023\\
125	0.161559790675023\\
126	0.161559790675023\\
127	0.161559790675023\\
128	0.161559790675023\\
129	0.161559790675023\\
130	0.161559790675023\\
131	0.161559790675023\\
132	0.161559790675023\\
133	0.161559790675023\\
134	0.161559790675023\\
135	0.161559790675023\\
136	0.161559790675023\\
137	0.161559790675023\\
138	0.161559790675023\\
139	0.161559790675023\\
140	0.161559790675023\\
141	0.161559790675023\\
142	0.161559790675023\\
143	0.161559790675023\\
144	0.161559790675023\\
145	0.161559790675023\\
146	0.161559790675023\\
147	0.161559790675023\\
148	0.161559790675023\\
149	0.161559790675023\\
150	0.161559790675023\\
151	0.161559790675023\\
152	0.161559790675023\\
153	0.161559790675023\\
154	0.161559790675023\\
155	0.161559790675023\\
156	0.161559790675023\\
157	0.161559790675023\\
158	0.161559790675023\\
159	0.161559790675023\\
160	0.161559790675023\\
161	0.161559790675023\\
162	0.161559790675023\\
163	0.161559790675023\\
164	0.161559790675023\\
165	0.161559790675023\\
166	0.161559790675023\\
167	0.161559790675023\\
168	0.161559790675023\\
169	0.161559790675023\\
170	0.161559790675023\\
171	0.161559790675023\\
172	0.161559790675023\\
173	0.161559790675023\\
174	0.161559790675023\\
175	0.161559790675023\\
176	0.161559790675023\\
177	0.161559790675023\\
178	0.161559790675023\\
179	0.161559790675023\\
180	0.161559790675023\\
181	0.161559790675023\\
182	0.161559790675023\\
183	0.161559790675023\\
184	0.161559790675023\\
185	0.161559790675023\\
186	0.161559790675023\\
187	0.161559790675023\\
188	0.161559790675023\\
189	0.161559790675023\\
190	0.161559790675023\\
191	0.161559790675023\\
192	0.161559790675023\\
193	0.161559790675023\\
194	0.161559790675023\\
195	0.161559790675023\\
196	0.161559790675023\\
197	0.161559790675023\\
198	0.161559790675023\\
199	0.161559790675023\\
200	0.161559790675023\\
};
\addlegendentry{Prop $(\alpha^{(k)} = 0.01/2^k)$}

\addplot [color=red, dashed, line width=1.2pt]
  table[row sep=crcr]{%
1	0.576582433691381\\
2	0.876381750605703\\
3	0.807941225499191\\
4	0.644073830050403\\
5	0.246516239760048\\
6	0.364621249866361\\
7	0.522137037132576\\
8	0.185621374721407\\
9	0.19987314582809\\
10	0.24510714704402\\
11	0.210609096072882\\
12	0.247754853284951\\
13	0.194413651893319\\
14	0.211630768978596\\
15	0.18668645003269\\
16	0.195629658311968\\
17	0.179752154984044\\
18	0.183995393778043\\
19	0.174258026991129\\
20	0.176054078252392\\
21	0.170124520555567\\
22	0.170726040449333\\
23	0.167114852703859\\
24	0.16717953926168\\
25	0.164972033410072\\
26	0.164827640609042\\
27	0.163470063607166\\
28	0.163269850566378\\
29	0.162428651571581\\
30	0.162237544049327\\
31	0.161711792996933\\
32	0.161552210520692\\
33	0.16122045961262\\
34	0.161095797846019\\
35	0.160884235354147\\
36	0.160790411890215\\
37	0.160653869792948\\
38	0.160584694629041\\
39	0.160495323169965\\
40	0.160444781837652\\
41	0.160385274602382\\
42	0.160348333812621\\
43	0.160307852953103\\
44	0.160280604320354\\
45	0.160252312771371\\
46	0.160231853493229\\
47	0.160211408657307\\
48	0.160195644662101\\
49	0.160180271309122\\
50	0.160167722232921\\
51	0.16015563710033\\
52	0.160145271685915\\
53	0.160135323446318\\
54	0.160126431121882\\
55	0.160117873376268\\
56	0.160109968588314\\
57	0.16010231572493\\
58	0.160095068782927\\
59	0.160088003874337\\
60	0.160081191992427\\
61	0.160074507617274\\
62	0.160067980775073\\
63	0.160061540813\\
64	0.160055198218372\\
65	0.160048913085312\\
66	0.160042687087398\\
67	0.160036497628906\\
68	0.160030342797883\\
69	0.160024209784627\\
70	0.160018095538881\\
71	0.160011992800495\\
72	0.160005898450499\\
73	0.159999808379643\\
74	0.159993719808183\\
75	0.159987630412493\\
76	0.15998153785741\\
77	0.159975440824318\\
78	0.159969337401097\\
79	0.159963226826438\\
80	0.159957107545672\\
81	0.159950979097809\\
82	0.159944840212754\\
83	0.159938690582694\\
84	0.159932529156497\\
85	0.159926355695913\\
86	0.159920169314748\\
87	0.159913969797511\\
88	0.159907756380452\\
89	0.159901528845351\\
90	0.1598952865183\\
91	0.159889029165128\\
92	0.159882756177087\\
93	0.159876467297802\\
94	0.159870161965099\\
95	0.159863839898027\\
96	0.159857500566985\\
97	0.159851143666134\\
98	0.159844768687872\\
99	0.159838375302175\\
100	0.15983196301539\\
101	0.159825531474426\\
102	0.159819080193401\\
103	0.159812608797362\\
104	0.159806116803389\\
105	0.159799603815792\\
106	0.159793069350827\\
107	0.159786512993009\\
108	0.159779934254744\\
109	0.159773332701498\\
110	0.159766707839355\\
111	0.159760059215273\\
112	0.159753386326961\\
113	0.159746688703212\\
114	0.159739965831626\\
115	0.159733217222993\\
116	0.159726442353295\\
117	0.159719640715331\\
118	0.159712811772127\\
119	0.159705954998361\\
120	0.159699069842893\\
121	0.159692155762025\\
122	0.159685212189333\\
123	0.159678238562384\\
124	0.159671234298414\\
125	0.159664198815801\\
126	0.159657131514428\\
127	0.159650031792947\\
128	0.159642899032898\\
129	0.159635732612598\\
130	0.159628531894267\\
131	0.159621296235208\\
132	0.159614024977346\\
133	0.159606717456228\\
134	0.159599372992501\\
135	0.159591990899156\\
136	0.159584570474565\\
137	0.159577111008311\\
138	0.159569611775471\\
139	0.159562072041308\\
140	0.159554491056564\\
141	0.159546868061224\\
142	0.159539202280617\\
143	0.159531492928439\\
144	0.159523739203504\\
145	0.159515940292154\\
146	0.159508095365541\\
147	0.159500203581544\\
148	0.159492264082469\\
149	0.159484275996566\\
150	0.159476238436073\\
151	0.159468150498392\\
152	0.15946001126442\\
153	0.159451819799449\\
154	0.159443575151713\\
155	0.159435276353072\\
156	0.159426922417733\\
157	0.159418512342746\\
158	0.159410045106873\\
159	0.159401519670923\\
160	0.159392934976742\\
161	0.159384289947413\\
162	0.159375583486337\\
163	0.159366814477328\\
164	0.159357981783761\\
165	0.159349084248576\\
166	0.159340120693484\\
167	0.15933108991889\\
168	0.159321990703144\\
169	0.159312821802392\\
170	0.159303581949866\\
171	0.159294269855671\\
172	0.159284884206103\\
173	0.159275423663384\\
174	0.159265886864994\\
175	0.159256272423366\\
176	0.159246578925227\\
177	0.159236804931252\\
178	0.159226948975416\\
179	0.159217009564607\\
180	0.159206985177987\\
181	0.159196874266569\\
182	0.15918667525258\\
183	0.159176386529011\\
184	0.159166006458974\\
185	0.15915553337523\\
186	0.159144965579542\\
187	0.159134301342176\\
188	0.159123538901254\\
189	0.159112676462232\\
190	0.159101712197249\\
191	0.159090644244584\\
192	0.159079470708\\
193	0.159068189656186\\
194	0.159056799122094\\
195	0.159045297102368\\
196	0.159033681556676\\
197	0.159021950407126\\
198	0.159010101537595\\
199	0.158998132793134\\
200	0.158986041979298\\
};
\addlegendentry{ADAGRAD $(\alpha^{(0)}= 1)$}

\end{axis}
\end{tikzpicture}%

%% file: figs/BF_train_batch_vs_stock.tex
\begin{tikzpicture}

\begin{axis}[%
width=0.85\columnwidth,
height=0.7\columnwidth,
at={(1.037in,0.517in)},
scale only axis,
xmin=1,
xmax=100,
xlabel={Iterations},
ymin=0.1,
ymax=0.6,
yminorticks=true,
xtick={1,20,40,60,80,100},
ytick={0.1,0.2,0.3,0.4,0.5,0.6},
ylabel={Normalized MSE},
ylabel near ticks,
xmajorgrids,
ymajorgrids,
axis background/.style={fill=white},
legend columns=1,
legend style={legend pos=north east,legend cell align=left,align=left,draw=black}
]
\addplot [color=blue, line width=1.2pt]
  table[row sep=crcr]{%
1	0.576582433691381\\
2	0.550564732104159\\
3	0.22419889030256\\
4	0.209094194358271\\
5	0.246431718643538\\
6	0.203838132071617\\
7	0.231613858497913\\
8	0.198242918104625\\
9	0.21803230351955\\
10	0.192498977691283\\
11	0.205822213048274\\
12	0.186776448783835\\
13	0.1950576747057\\
14	0.181262611680581\\
15	0.185812078352353\\
16	0.176154330740969\\
17	0.178142977274781\\
18	0.171644571345731\\
19	0.172070037891538\\
20	0.167897264444067\\
21	0.167545230104265\\
22	0.165011796807147\\
23	0.164426259682356\\
24	0.162989186571348\\
25	0.162472589679403\\
26	0.161721857681252\\
27	0.161379132348964\\
28	0.161023428288172\\
29	0.160839578609784\\
30	0.16068872244867\\
31	0.160606629663208\\
32	0.160549705168571\\
33	0.160518408358629\\
34	0.160499028204812\\
35	0.160488290715308\\
36	0.160481874265028\\
37	0.160478032408148\\
38	0.160475469591093\\
39	0.160473577353175\\
40	0.160472006601338\\
41	0.160470591855088\\
42	0.16046925137319\\
43	0.160467948092551\\
44	0.160466665573199\\
45	0.160465396761775\\
46	0.160464138763426\\
47	0.160462890348225\\
48	0.16046165099476\\
49	0.160460420428302\\
50	0.160459198480366\\
51	0.160457985013538\\
52	0.160456779906647\\
53	0.16045558304317\\
54	0.160454394310735\\
55	0.160453213599208\\
56	0.160452040800931\\
57	0.160450875810327\\
58	0.16044971852396\\
59	0.160448568840408\\
60	0.160447426660238\\
61	0.160446291885946\\
62	0.160445164421908\\
63	0.160444044174332\\
64	0.160442931051213\\
65	0.160441824962287\\
66	0.160440725818989\\
67	0.160439633534406\\
68	0.160438548023243\\
69	0.160437469201777\\
70	0.160436396987825\\
71	0.160435331300701\\
72	0.160434272061181\\
73	0.160433219191471\\
74	0.160432172615172\\
75	0.160431132257242\\
76	0.160430098043973\\
77	0.16042906990295\\
78	0.160428047763029\\
79	0.160427031554303\\
80	0.160426021208075\\
81	0.160425016656829\\
82	0.160424017834207\\
83	0.160423024674978\\
84	0.160422037115014\\
85	0.160421055091268\\
86	0.160420078541746\\
87	0.160419107405488\\
88	0.160418141622542\\
89	0.160417181133941\\
90	0.160416225881685\\
91	0.160415275808719\\
92	0.160414330858912\\
93	0.160413390977036\\
94	0.160412456108748\\
95	0.160411526200574\\
96	0.160410601199885\\
97	0.160409681054884\\
98	0.160408765714586\\
99	0.160407855128802\\
100	0.160406949248124\\
101	0.160406048023906\\
102	0.160405151408251\\
103	0.160404259353993\\
104	0.160403371814686\\
105	0.160402488744587\\
106	0.160401610098642\\
107	0.160400735832471\\
108	0.16039986590236\\
109	0.16039900026524\\
110	0.160398138878682\\
111	0.160397281700878\\
112	0.160396428690633\\
113	0.160395579807352\\
114	0.160394735011029\\
115	0.160393894262234\\
116	0.160393057522103\\
117	0.160392224752328\\
118	0.160391395915146\\
119	0.160390570973327\\
120	0.160389749890168\\
121	0.160388932629479\\
122	0.160388119155575\\
123	0.160387309433268\\
124	0.160386503427856\\
125	0.160385701105115\\
126	0.160384902431289\\
127	0.160384107373086\\
128	0.160383315897662\\
129	0.16038252797262\\
130	0.160381743565998\\
131	0.160380962646263\\
132	0.160380185182302\\
133	0.160379411143416\\
134	0.160378640499313\\
135	0.160377873220097\\
136	0.160377109276268\\
137	0.160376348638707\\
138	0.160375591278678\\
139	0.160374837167814\\
140	0.160374086278113\\
141	0.160373338581936\\
142	0.160372594051994\\
143	0.160371852661347\\
144	0.160371114383395\\
145	0.160370379191875\\
146	0.160369647060855\\
147	0.160368917964725\\
148	0.160368191878196\\
149	0.160367468776292\\
150	0.160366748634348\\
151	0.160366031428\\
152	0.160365317133184\\
153	0.160364605726131\\
154	0.160363897183358\\
155	0.16036319148167\\
156	0.160362488598151\\
157	0.160361788510158\\
158	0.160361091195323\\
159	0.160360396631542\\
160	0.160359704796975\\
161	0.160359015670039\\
162	0.160358329229407\\
163	0.160357645454001\\
164	0.160356964322991\\
165	0.160356285815788\\
166	0.160355609912043\\
167	0.160354936591642\\
168	0.160354265834704\\
169	0.160353597621573\\
170	0.160352931932821\\
171	0.160352268749238\\
172	0.160351608051836\\
173	0.160350949821838\\
174	0.160350294040679\\
175	0.160349640690005\\
176	0.160348989751664\\
177	0.160348341207707\\
178	0.160347695040386\\
179	0.160347051232145\\
180	0.160346409765627\\
181	0.160345770623661\\
182	0.160345133789265\\
183	0.160344499245643\\
184	0.160343866976181\\
185	0.160343236964444\\
186	0.160342609194174\\
187	0.160341983649289\\
188	0.160341360313878\\
189	0.1603407391722\\
190	0.160340120208681\\
191	0.160339503407913\\
192	0.160338888754649\\
193	0.160338276233804\\
194	0.160337665830449\\
195	0.160337057529814\\
196	0.16033645131728\\
197	0.16033584717838\\
198	0.160335245098798\\
199	0.160334645064364\\
200	0.160334047061054\\
};
\addlegendentry{Prop (setting 1)}

\addplot [color=black, dashed, line width=1.2pt]
  table[row sep=crcr]{%
1	0.589827484116837\\
2	0.199489842622258\\
3	0.31297374332088\\
4	0.132310776223076\\
5	0.159113985209024\\
6	0.128345456186279\\
7	0.135525645817761\\
8	0.160741452441902\\
9	0.156537340363362\\
10	0.172579173192933\\
11	0.257399456283527\\
12	0.160120181816653\\
13	0.146984586812192\\
14	0.136434243960527\\
15	0.201857879531121\\
16	0.162249931752907\\
17	0.18158478613801\\
18	0.187685530771426\\
19	0.146379781330208\\
20	0.139279636820436\\
21	0.133275158074475\\
22	0.134233613605776\\
23	0.0924095806546597\\
24	0.157451262455258\\
25	0.165966338666121\\
26	0.128272957472427\\
27	0.142464834762987\\
28	0.153103298827887\\
29	0.157368532713402\\
30	0.156284503107809\\
31	0.143225759300165\\
32	0.166602123953567\\
33	0.143359603079718\\
34	0.179419214674066\\
35	0.156300067562498\\
36	0.182172632279618\\
37	0.174406118344696\\
38	0.181222375965754\\
39	0.0996706808441733\\
40	0.148287199119843\\
41	0.174763811408626\\
42	0.160305304589873\\
43	0.18136045947622\\
44	0.167574643820172\\
45	0.166287475450612\\
46	0.171772452470658\\
47	0.148066718350092\\
48	0.17854111672252\\
49	0.225471960125091\\
50	0.165398988660923\\
51	0.170266615991662\\
52	0.187337361455888\\
53	0.137795870285347\\
54	0.157942361617756\\
55	0.145996699074471\\
56	0.140316335564918\\
57	0.160715639117479\\
58	0.145255504305381\\
59	0.163819228990958\\
60	0.162940695248404\\
61	0.157295992589942\\
62	0.126568376519844\\
63	0.140817206223009\\
64	0.130555210614546\\
65	0.135215839775201\\
66	0.144772830954302\\
67	0.131292895437942\\
68	0.133578469918317\\
69	0.172238743080054\\
70	0.171568744955192\\
71	0.207318707043581\\
72	0.191932594278939\\
73	0.208657900729625\\
74	0.186104354791611\\
75	0.142655370042082\\
76	0.169968714807256\\
77	0.160352636018425\\
78	0.147494405477724\\
79	0.119258269410631\\
80	0.203622495349111\\
81	0.17657215731805\\
82	0.218280258547049\\
83	0.191939304570238\\
84	0.160007714323187\\
85	0.186899636329627\\
86	0.154798907100747\\
87	0.182022155087778\\
88	0.180322218976061\\
89	0.217108281125289\\
90	0.170538774252389\\
91	0.168124426246374\\
92	0.163508241367253\\
93	0.148621155366092\\
94	0.233167461962551\\
95	0.154828845421189\\
96	0.148353942859752\\
97	0.1595493790083\\
98	0.19191868680102\\
99	0.177164461417852\\
100	0.157579351774406\\
101	0.192930653749137\\
102	0.168950289876787\\
103	0.186999182372956\\
104	0.182770145259488\\
105	0.157746188766934\\
106	0.168109842721149\\
107	0.177542881654225\\
108	0.144569746625693\\
109	0.140590116983123\\
110	0.102992705044235\\
111	0.184734616082274\\
112	0.150308359033529\\
113	0.177941744091837\\
114	0.134865615174753\\
115	0.159285133503334\\
116	0.0956122271148674\\
117	0.138876305321829\\
118	0.198043039359114\\
119	0.166740467762449\\
120	0.182833953080526\\
121	0.136437639618877\\
122	0.165488661632106\\
123	0.177249445665465\\
124	0.153460309422034\\
125	0.13552750079999\\
126	0.109685639009233\\
127	0.148974417782438\\
128	0.134426175566116\\
129	0.110963810985105\\
130	0.159955861905439\\
131	0.185661497393621\\
132	0.194269656086808\\
133	0.181530133883311\\
134	0.139339979873132\\
135	0.142013612719122\\
136	0.179421829235826\\
137	0.198618576398851\\
138	0.174904518955699\\
139	0.176924628209293\\
140	0.161438909270845\\
141	0.173782414520557\\
142	0.162353887736965\\
143	0.157422740996261\\
144	0.170149117993195\\
145	0.144546502933457\\
146	0.191287237713382\\
147	0.165186008365866\\
148	0.173824582477694\\
149	0.153695256211042\\
150	0.152294043314351\\
151	0.143461734554584\\
152	0.142394010199055\\
153	0.21845558426034\\
154	0.175682762805901\\
155	0.209062603252894\\
156	0.140276973144744\\
157	0.186481528640027\\
158	0.172324669088011\\
159	0.153451687974674\\
160	0.148688629686567\\
161	0.142556330017574\\
162	0.126083696085777\\
163	0.144839698518101\\
164	0.180510525212537\\
165	0.148150735421904\\
166	0.193384271111352\\
167	0.205651933293867\\
168	0.150552789740078\\
169	0.165560686101583\\
170	0.163986423609546\\
171	0.151323732295126\\
172	0.172738563835061\\
173	0.134870173033221\\
174	0.167279550933135\\
175	0.168778222771264\\
176	0.179442220787012\\
177	0.140502201057485\\
178	0.190128561087369\\
179	0.169597442662757\\
180	0.169965924513515\\
181	0.132260222589097\\
182	0.201775903138886\\
183	0.193185664755829\\
184	0.173308537383504\\
185	0.149929660817491\\
186	0.168867017181224\\
187	0.121341066930952\\
188	0.122732926227889\\
189	0.139643400579926\\
190	0.181830775644315\\
191	0.131416316085934\\
192	0.16930649708445\\
193	0.195231908226885\\
194	0.168093004016304\\
195	0.200271751897368\\
196	0.183242563049432\\
197	0.103664793257258\\
198	0.210748155830187\\
199	0.158352204886029\\
200	0.137232010412453\\
};
\addlegendentry{Prop (Stochastic, $B = 50$)}

\addplot [color=red, dashed, line width=1.2pt]
  table[row sep=crcr]{%
1	0.587428000814008\\
2	0.157142649066353\\
3	0.146354792701031\\
4	0.164217006783509\\
5	0.168801686371912\\
6	0.128730602549389\\
7	0.139656173870317\\
8	0.149067618103822\\
9	0.117008387834231\\
10	0.160709400789589\\
11	0.117230600034824\\
12	0.207468129598131\\
13	0.174384726599952\\
14	0.177746104100471\\
15	0.143883979065683\\
16	0.164556849262938\\
17	0.118431390206921\\
18	0.159284149324493\\
19	0.135288065103525\\
20	0.119703105964892\\
21	0.195753160177049\\
22	0.212986127227053\\
23	0.155009858656252\\
24	0.167571484126668\\
25	0.17981822865483\\
26	0.149202131970488\\
27	0.178595186015451\\
28	0.160862208067452\\
29	0.127463627865761\\
30	0.130175086037581\\
31	0.133716615236142\\
32	0.160883917362322\\
33	0.156391836152031\\
34	0.177612507526809\\
35	0.141815086011208\\
36	0.197332949500398\\
37	0.142891962915084\\
38	0.134298841811797\\
39	0.145876073046064\\
40	0.173557681393814\\
41	0.144883010767137\\
42	0.219500494775677\\
43	0.116531632000288\\
44	0.145797004008924\\
45	0.184209018301862\\
46	0.180340635752367\\
47	0.20263444203663\\
48	0.166474730867024\\
49	0.151642608076012\\
50	0.16784782351964\\
51	0.2022813858742\\
52	0.163613669785099\\
53	0.169278862199568\\
54	0.178831761242788\\
55	0.175070493709274\\
56	0.132633937815653\\
57	0.186444227000179\\
58	0.148625172723533\\
59	0.160476502939179\\
60	0.180542294195331\\
61	0.146720701026405\\
62	0.16020138716032\\
63	0.201334640386619\\
64	0.154578574924918\\
65	0.14259137820518\\
66	0.185022199488249\\
67	0.172775396152593\\
68	0.186349505817293\\
69	0.180765905188107\\
70	0.156395445227637\\
71	0.169094266085115\\
72	0.164437671392744\\
73	0.143590335606979\\
74	0.172217000664661\\
75	0.177990893501223\\
76	0.164679484470519\\
77	0.145784039393937\\
78	0.142223000472568\\
79	0.156104493313726\\
80	0.186653269639383\\
81	0.170007889797665\\
82	0.15740689314937\\
83	0.151280827267325\\
84	0.171554925628217\\
85	0.175095611576281\\
86	0.178045572307422\\
87	0.163703477071035\\
88	0.155825163087359\\
89	0.13964395074909\\
90	0.155746075024604\\
91	0.181399012048457\\
92	0.179094852976262\\
93	0.114582298213319\\
94	0.169882815452254\\
95	0.172063647736707\\
96	0.135913798608667\\
97	0.166487479905995\\
98	0.182497851730191\\
99	0.160640173797445\\
100	0.125864603766421\\
101	0.160826401718439\\
102	0.138736588514429\\
103	0.165167313288872\\
104	0.139028647163663\\
105	0.161339933653072\\
106	0.150967868440492\\
107	0.157135410454899\\
108	0.144378189809099\\
109	0.159460501575296\\
110	0.183939075237154\\
111	0.151142959014739\\
112	0.169194137396029\\
113	0.13510594686489\\
114	0.15117089922615\\
115	0.144593989641236\\
116	0.152467709620297\\
117	0.138773105039076\\
118	0.142032648094912\\
119	0.18626626065328\\
120	0.138186039994076\\
121	0.198003999098148\\
122	0.155989465498741\\
123	0.175157219542088\\
124	0.164413277699209\\
125	0.174728935746306\\
126	0.150464564261035\\
127	0.193078466309713\\
128	0.147460723501606\\
129	0.135745606410187\\
130	0.140107723992732\\
131	0.213317497979717\\
132	0.171312347581487\\
133	0.146752270577438\\
134	0.135757604031059\\
135	0.123243959476554\\
136	0.174219619443043\\
137	0.167249836824652\\
138	0.190364636237059\\
139	0.157537752698301\\
140	0.150501464143993\\
141	0.151956108055833\\
142	0.166626413348774\\
143	0.158153080749449\\
144	0.115179590552567\\
145	0.213173320221732\\
146	0.150186332002078\\
147	0.13893351637389\\
148	0.17371569714404\\
149	0.161380102239418\\
150	0.137209256889611\\
151	0.117646695626296\\
152	0.137734743552131\\
153	0.146608766021979\\
154	0.163246881937664\\
155	0.14758803939815\\
156	0.15374591212667\\
157	0.17595562486554\\
158	0.165634235653619\\
159	0.136193033755843\\
160	0.155178225206317\\
161	0.14073368699888\\
162	0.16407825525089\\
163	0.149359434997216\\
164	0.170495191809715\\
165	0.18087379021804\\
166	0.163588998030287\\
167	0.124824394844887\\
168	0.17179741930423\\
169	0.166501623830643\\
170	0.139632220393713\\
171	0.182939488964317\\
172	0.186507920430115\\
173	0.156370940621484\\
174	0.160464965389486\\
175	0.118257736817886\\
176	0.126866868911499\\
177	0.138773696397192\\
178	0.142002154859415\\
179	0.199622780637708\\
180	0.143088685203644\\
181	0.161734818900088\\
182	0.181408499321367\\
183	0.161010241659601\\
184	0.144923488501628\\
185	0.203854395174096\\
186	0.143914355983391\\
187	0.150927147661974\\
188	0.169860665150419\\
189	0.192554431851196\\
190	0.17127091576017\\
191	0.14032112241597\\
192	0.153106153510266\\
193	0.179798964301791\\
194	0.134097310652999\\
195	0.153596034056591\\
196	0.177477339553316\\
197	0.157621820123\\
198	0.174307799750958\\
199	0.15819145078122\\
200	0.130012977520835\\
};
\addlegendentry{Prop (Stochastic, $B = 200$)}

\addplot [color=green, dashed, line width=1.2pt]
  table[row sep=crcr]{%
1	0.558585229345778\\
2	0.370451148325549\\
3	0.173290857467582\\
4	0.15430321311174\\
5	0.149409753282428\\
6	0.16127557352193\\
7	0.166295896963023\\
8	0.17557904840505\\
9	0.162320600093308\\
10	0.165586087682977\\
11	0.15359416086656\\
12	0.196204665062045\\
13	0.17941371750111\\
14	0.166866919957056\\
15	0.167749623050406\\
16	0.195705371931853\\
17	0.165259364566645\\
18	0.14798719445307\\
19	0.139754664001394\\
20	0.154800753543953\\
21	0.159787360851457\\
22	0.164200430498438\\
23	0.167948336771543\\
24	0.165407080094297\\
25	0.165296887283514\\
26	0.163841109472365\\
27	0.163585192817858\\
28	0.145001227686264\\
29	0.190296508421329\\
30	0.179549820237606\\
31	0.161907116720806\\
32	0.165495989887346\\
33	0.187709629337876\\
34	0.188815830851643\\
35	0.189585111986579\\
36	0.16964915267311\\
37	0.148405350401543\\
38	0.163583144637717\\
39	0.169068182832866\\
40	0.168858787895542\\
41	0.15036621870028\\
42	0.163148879920419\\
43	0.151246659028483\\
44	0.138167007264512\\
45	0.160494263675337\\
46	0.148773690026728\\
47	0.15257560651714\\
48	0.148564239190725\\
49	0.155510532675374\\
50	0.159972804724485\\
51	0.160842807349981\\
52	0.169438214667897\\
53	0.180330109850871\\
54	0.148979380878875\\
55	0.170119565008025\\
56	0.150691298097103\\
57	0.152005909428329\\
58	0.180697241884146\\
59	0.158193505180181\\
60	0.161265142020718\\
61	0.161526597502493\\
62	0.168365476059857\\
63	0.174281941710647\\
64	0.166296877977645\\
65	0.153368477509657\\
66	0.163350271323136\\
67	0.198522427940821\\
68	0.17585731510327\\
69	0.16152164604685\\
70	0.185772220285017\\
71	0.186799641071234\\
72	0.169508105179282\\
73	0.167283293033427\\
74	0.181355861049188\\
75	0.160127182195638\\
76	0.178480339131827\\
77	0.159973046402151\\
78	0.180170427797291\\
79	0.15714518319703\\
80	0.146961824850496\\
81	0.17612351372756\\
82	0.185215895205613\\
83	0.14909757402029\\
84	0.154705110394045\\
85	0.164025241135834\\
86	0.150558660425686\\
87	0.17281834249943\\
88	0.159744122294295\\
89	0.168041754027492\\
90	0.159786553184919\\
91	0.157633730202382\\
92	0.158161049996876\\
93	0.161401433148925\\
94	0.170819248506184\\
95	0.191435784932525\\
96	0.139941272638076\\
97	0.173168781493808\\
98	0.158802696651466\\
99	0.167189808682622\\
100	0.15604241796565\\
101	0.143820908291721\\
102	0.14259525566266\\
103	0.164518738124531\\
104	0.160611060177785\\
105	0.198997116544719\\
106	0.171750278631996\\
107	0.145746717865095\\
108	0.14820482449913\\
109	0.162865924919108\\
110	0.133521669389909\\
111	0.160828846175185\\
112	0.180473837454903\\
113	0.164903051118688\\
114	0.150389253664833\\
115	0.17987721725316\\
116	0.137730613355931\\
117	0.134046516043\\
118	0.170226984657368\\
119	0.17115495555059\\
120	0.185956883359143\\
121	0.162286612587868\\
122	0.185899100303073\\
123	0.151892143168652\\
124	0.150062649501002\\
125	0.14046995133898\\
126	0.175851115576765\\
127	0.175957778064142\\
128	0.141708456458998\\
129	0.18143364923959\\
130	0.158336001313878\\
131	0.158690199341869\\
132	0.150839872551293\\
133	0.17279272988092\\
134	0.184336011379925\\
135	0.141130594944779\\
136	0.168061494115904\\
137	0.168614346612348\\
138	0.158497055638097\\
139	0.15786098088208\\
140	0.149198253160225\\
141	0.177550290534779\\
142	0.156482382909659\\
143	0.163064859284166\\
144	0.16151164752795\\
145	0.185498355315399\\
146	0.15422064350269\\
147	0.163523105296922\\
148	0.146566323592591\\
149	0.163758288429117\\
150	0.139585796595579\\
151	0.137417791405151\\
152	0.165249179239976\\
153	0.149141395888827\\
154	0.15834508359728\\
155	0.157945246559157\\
156	0.154703195010418\\
157	0.145711055246448\\
158	0.163486525797231\\
159	0.142509053272847\\
160	0.145435842294074\\
161	0.157445238058708\\
162	0.169436428793952\\
163	0.18591006825771\\
164	0.171697019797683\\
165	0.174239573685499\\
166	0.160576767302916\\
167	0.17160163892676\\
168	0.165287984517749\\
169	0.156846003367094\\
170	0.156074839086462\\
171	0.159330925003837\\
172	0.164099695550275\\
173	0.169073445906608\\
174	0.195315529755403\\
175	0.153232705127787\\
176	0.135665228929229\\
177	0.175122121290663\\
178	0.168573033097289\\
179	0.146936989244727\\
180	0.187709678240286\\
181	0.155386965910442\\
182	0.141347069205101\\
183	0.148815862332261\\
184	0.157589147228306\\
185	0.135996026137079\\
186	0.174409495171333\\
187	0.146461052381787\\
188	0.189915615720992\\
189	0.167981575241459\\
190	0.164835403046682\\
191	0.166744689807372\\
192	0.160631373387721\\
193	0.146800354758504\\
194	0.135455765239017\\
195	0.197521817174963\\
196	0.157637322969049\\
197	0.194750443030311\\
198	0.153256736610278\\
199	0.169167083813514\\
200	0.147058851706934\\
};
\addlegendentry{Prop (Stochastic, increasing $B$)}

\end{axis}
\end{tikzpicture}%

%% file: figs/BF_training_convexblck1.tex
\begin{tikzpicture}

\begin{axis}[%
width=0.8\columnwidth,
height=0.4\columnwidth,
at={(1.037in,0.517in)},
scale only axis,
xmin=1,
xmax=200,
xlabel={Iterations},
ymode=log,
ymin=1,
ymax=3.1,
yminorticks=true,
xtick={1,50,100,150,200},
ytick={1.2,1.4,1.6,1.8,2,2.2,2.4,2.6,2.8,3},
ylabel={Normalized MSE},
xmajorgrids,
ymajorgrids,
log ticks with fixed point,
axis background/.style={fill=white},
legend columns=1,
legend style={legend pos=north east,legend cell align=left,align=left,draw=black}
]
\addplot [color=blue, line width=1.2pt]
  table[row sep=crcr]{%
1	3.0481117080741\\
2	1.62653320831785\\
3	1.62651606479312\\
4	1.62649863295613\\
5	1.62648090578612\\
6	1.62646287603807\\
7	1.62644453623381\\
8	1.6264258786527\\
9	1.62640689532191\\
10	1.62638757800621\\
11	1.6263679181972\\
12	1.62634790710214\\
13	1.62632753563214\\
14	1.62630679438978\\
15	1.62628567365613\\
16	1.6262641633771\\
17	1.62624225314912\\
18	1.62621993220404\\
19	1.62619718939332\\
20	1.62617401317134\\
21	1.62615039157784\\
22	1.6261263122195\\
23	1.62610176225041\\
24	1.62607672835165\\
25	1.6260511967096\\
26	1.62602515299322\\
27	1.62599858232993\\
28	1.62597146928028\\
29	1.6259437978111\\
30	1.62591555126719\\
31	1.62588671234135\\
32	1.62585726304274\\
33	1.62582718466335\\
34	1.62579645774251\\
35	1.62576506202933\\
36	1.62573297644286\\
37	1.62570017902985\\
38	1.62566664691999\\
39	1.62563235627831\\
40	1.62559728225469\\
41	1.62556139893018\\
42	1.62552467925997\\
43	1.62548709501261\\
44	1.6254486167054\\
45	1.62540921353553\\
46	1.62536885330669\\
47	1.62532750235081\\
48	1.62528512544444\\
49	1.62524168571962\\
50	1.62519714456844\\
51	1.62515146154112\\
52	1.62510459423681\\
53	1.62505649818675\\
54	1.62500712672896\\
55	1.62495643087388\\
56	1.62490435916013\\
57	1.62485085749955\\
58	1.62479586901062\\
59	1.62473933383922\\
60	1.62468118896557\\
61	1.62462136799618\\
62	1.62455980093939\\
63	1.62449641396302\\
64	1.62443112913242\\
65	1.62436386412717\\
66	1.62429453193418\\
67	1.62422304051524\\
68	1.62414929244611\\
69	1.62407318452468\\
70	1.62399460734484\\
71	1.62391344483274\\
72	1.62382957374133\\
73	1.62374286309906\\
74	1.62365317360765\\
75	1.62356035698355\\
76	1.62346425523689\\
77	1.62336469988093\\
78	1.62326151106433\\
79	1.62315449661712\\
80	1.62304345100069\\
81	1.62292815415018\\
82	1.62280837019651\\
83	1.62268384605344\\
84	1.62255430985271\\
85	1.62241946920853\\
86	1.62227900928913\\
87	1.62213259067069\\
88	1.62197984694452\\
89	1.62182038204443\\
90	1.62165376725585\\
91	1.62147953786224\\
92	1.62129718937727\\
93	1.62110617330269\\
94	1.62090589234188\\
95	1.62069569498706\\
96	1.62047486938408\\
97	1.62024263636142\\
98	1.61999814149004\\
99	1.61974044601541\\
100	1.61946851647409\\
101	1.61918121277024\\
102	1.61887727444401\\
103	1.61855530480942\\
104	1.61821375257324\\
105	1.6178508904647\\
106	1.61746479030436\\
107	1.6170532938144\\
108	1.61661397831484\\
109	1.61614411625145\\
110	1.61564062725049\\
111	1.61510002107618\\
112	1.61451832945966\\
113	1.61389102424345\\
114	1.61321291860797\\
115	1.61247804726351\\
116	1.61167952033394\\
117	1.6108093441313\\
118	1.60985819998966\\
119	1.60881516960747\\
120	1.60766739167584\\
121	1.60639962957486\\
122	1.60499372306547\\
123	1.60342788741956\\
124	1.60167581019337\\
125	1.59970547721418\\
126	1.59747763289329\\
127	1.59494374211166\\
128	1.5920432663288\\
129	1.58869998738071\\
130	1.58481699717081\\
131	1.58026980375267\\
132	1.57489676209562\\
133	1.56848569500065\\
134	1.56075510609281\\
135	1.55132782343277\\
136	1.53969441432594\\
137	1.52516387109732\\
138	1.50680162318951\\
139	1.48336446087585\\
140	1.45326922183902\\
141	1.41469787697963\\
142	1.36606802756453\\
143	1.30723317052369\\
144	1.24156003108173\\
145	1.17764644522448\\
146	1.12720458563718\\
147	1.09743130211482\\
148	1.08521229823508\\
149	1.08183231155346\\
150	1.08117022877368\\
151	1.08106697045469\\
152	1.08105259985781\\
153	1.08105068802396\\
154	1.08105043536754\\
155	1.0810503997856\\
156	1.08105039247207\\
157	1.08105038884198\\
158	1.08105038569463\\
159	1.08105038261262\\
160	1.08105037954138\\
161	1.08105037647377\\
162	1.08105037340883\\
163	1.08105037034643\\
164	1.08105036728657\\
165	1.08105036422923\\
166	1.08105036117441\\
167	1.0810503581221\\
168	1.0810503550723\\
169	1.08105035202501\\
170	1.08105034898022\\
171	1.08105034593793\\
172	1.08105034289813\\
173	1.08105033986082\\
174	1.081050336826\\
175	1.08105033379366\\
176	1.0810503307638\\
177	1.08105032773641\\
178	1.0810503247115\\
179	1.08105032168905\\
180	1.08105031866906\\
181	1.08105031565153\\
182	1.08105031263645\\
183	1.08105030962382\\
184	1.08105030661364\\
185	1.08105030360591\\
186	1.08105030060061\\
187	1.08105029759775\\
188	1.08105029459732\\
189	1.08105029159931\\
190	1.08105028860373\\
191	1.08105028561057\\
192	1.08105028261982\\
193	1.08105027963149\\
194	1.08105027664556\\
195	1.08105027366204\\
196	1.08105027068092\\
197	1.0810502677022\\
198	1.08105026472587\\
199	1.08105026175192\\
200	1.08105025878037\\
};
\addlegendentry{Prop (setting 2)}

\addplot [color=green, line width=1.2pt]
  table[row sep=crcr]{%
1	3.0481117080741\\
2	1.6265712357933\\
3	1.62655956866961\\
4	1.62654993013714\\
5	1.62654150122214\\
6	1.62653389759219\\
7	1.62652690286559\\
8	1.62652038108002\\
9	1.62651424024503\\
10	1.62650841470718\\
11	1.62650285567662\\
12	1.62649752572762\\
13	1.62649239540892\\
14	1.62648744105211\\
15	1.6264826432981\\
16	1.62647798607357\\
17	1.62647345586016\\
18	1.62646904116017\\
19	1.62646473209774\\
20	1.62646052011592\\
21	1.62645639774269\\
22	1.62645235840785\\
23	1.62644839629775\\
24	1.62644450623882\\
25	1.62644068360314\\
26	1.62643692423123\\
27	1.6264332243683\\
28	1.62642958061124\\
29	1.62642598986426\\
30	1.6264224493014\\
31	1.62641895633476\\
32	1.62641550858737\\
33	1.62641210386994\\
34	1.62640874016074\\
35	1.62640541558822\\
36	1.62640212841586\\
37	1.62639887702896\\
38	1.62639565992296\\
39	1.62639247569328\\
40	1.62638932302625\\
41	1.62638620069107\\
42	1.62638310753272\\
43	1.62638004246559\\
44	1.62637700446776\\
45	1.62637399257591\\
46	1.62637100588077\\
47	1.62636804352289\\
48	1.62636510468901\\
49	1.62636218860858\\
50	1.62635929455072\\
51	1.62635642182143\\
52	1.62635356976106\\
53	1.62635073774192\\
54	1.62634792516621\\
55	1.62634513146407\\
56	1.62634235609175\\
57	1.62633959853\\
58	1.62633685828254\\
59	1.62633413487468\\
60	1.62633142785201\\
61	1.62632873677923\\
62	1.62632606123904\\
63	1.62632340083109\\
64	1.62632075517109\\
65	1.62631812388987\\
66	1.62631550663258\\
67	1.62631290305795\\
68	1.62631031283754\\
69	1.62630773565508\\
70	1.62630517120588\\
71	1.6263026191962\\
72	1.62630007934277\\
73	1.62629755137221\\
74	1.62629503502058\\
75	1.62629253003295\\
76	1.62629003616296\\
77	1.62628755317241\\
78	1.62628508083092\\
79	1.62628261891553\\
80	1.62628016721042\\
81	1.62627772550656\\
82	1.62627529360142\\
83	1.62627287129872\\
84	1.62627045840811\\
85	1.62626805474499\\
86	1.62626566013019\\
87	1.62626327438984\\
88	1.62626089735507\\
89	1.62625852886186\\
90	1.62625616875084\\
91	1.62625381686711\\
92	1.62625147306004\\
93	1.62624913718313\\
94	1.62624680909388\\
95	1.62624448865357\\
96	1.62624217572718\\
97	1.62623987018324\\
98	1.62623757189369\\
99	1.62623528073375\\
100	1.62623299658184\\
101	1.62623071931942\\
102	1.62622844883093\\
103	1.62622618500364\\
104	1.62622392772759\\
105	1.62622167689549\\
106	1.62621943240262\\
107	1.62621719414674\\
108	1.62621496202803\\
109	1.62621273594898\\
110	1.62621051581435\\
111	1.62620830153107\\
112	1.62620609300819\\
113	1.62620389015681\\
114	1.62620169288999\\
115	1.62619950112272\\
116	1.62619731477186\\
117	1.62619513375607\\
118	1.62619295799575\\
119	1.626190787413\\
120	1.62618862193159\\
121	1.62618646147687\\
122	1.62618430597573\\
123	1.62618215535659\\
124	1.62618000954933\\
125	1.62617786848524\\
126	1.62617573209702\\
127	1.62617360031868\\
128	1.62617147308555\\
129	1.62616935033425\\
130	1.62616723200261\\
131	1.62616511802967\\
132	1.62616300835564\\
133	1.62616090292188\\
134	1.62615880167083\\
135	1.62615670454605\\
136	1.62615461149212\\
137	1.62615252245464\\
138	1.62615043738024\\
139	1.6261483562165\\
140	1.62614627891194\\
141	1.62614420541602\\
142	1.62614213567909\\
143	1.62614006965238\\
144	1.62613800728798\\
145	1.62613594853881\\
146	1.62613389335862\\
147	1.62613184170193\\
148	1.62612979352406\\
149	1.62612774878107\\
150	1.62612570742976\\
151	1.62612366942767\\
152	1.62612163473304\\
153	1.62611960330477\\
154	1.62611757510248\\
155	1.62611555008642\\
156	1.62611352821748\\
157	1.62611150945718\\
158	1.62610949376768\\
159	1.6261074811117\\
160	1.62610547145257\\
161	1.6261034647542\\
162	1.62610146098104\\
163	1.62609946009811\\
164	1.62609746207096\\
165	1.62609546686565\\
166	1.62609347444879\\
167	1.62609148478745\\
168	1.62608949784923\\
169	1.62608751360219\\
170	1.62608553201487\\
171	1.62608355305628\\
172	1.62608157669586\\
173	1.62607960290353\\
174	1.62607763164961\\
175	1.62607566290485\\
176	1.62607369664045\\
177	1.62607173282799\\
178	1.62606977143944\\
179	1.6260678124472\\
180	1.62606585582401\\
181	1.62606390154304\\
182	1.62606194957777\\
183	1.6260599999021\\
184	1.62605805249025\\
185	1.6260561073168\\
186	1.62605416435667\\
187	1.62605222358512\\
188	1.62605028497773\\
189	1.62604834851042\\
190	1.62604641415941\\
191	1.62604448190123\\
192	1.62604255171274\\
193	1.62604062357108\\
194	1.62603869745368\\
195	1.62603677333827\\
196	1.62603485120287\\
197	1.62603293102575\\
198	1.62603101278549\\
199	1.62602909646092\\
200	1.62602718203114\\
};
\addlegendentry{Prop $(\alpha^{(k)}= 1/\sqrt{k})$}

\addplot [color=red, line width=1.2pt]
  table[row sep=crcr]{%
1	3.0481117080741\\
2	1.23579328349824\\
3	1.22344799011685\\
4	1.21738855975724\\
5	1.21439311613861\\
6	1.2129046872021\\
7	1.21216288483212\\
8	1.21179259765957\\
9	1.21160760894184\\
10	1.21151515347028\\
11	1.21146893547757\\
12	1.21144582891963\\
13	1.21143427625059\\
14	1.2114285000686\\
15	1.21142561201574\\
16	1.21142416799885\\
17	1.21142344599279\\
18	1.21142308499035\\
19	1.21142290448928\\
20	1.21142281423878\\
21	1.21142276911354\\
22	1.21142274655093\\
23	1.21142273526962\\
24	1.21142272962896\\
25	1.21142272680864\\
26	1.21142272539847\\
27	1.21142272469339\\
28	1.21142272434085\\
29	1.21142272416458\\
30	1.21142272407645\\
31	1.21142272403238\\
32	1.21142272401034\\
33	1.21142272399933\\
34	1.21142272399382\\
35	1.21142272399107\\
36	1.21142272398969\\
37	1.211422723989\\
38	1.21142272398866\\
39	1.21142272398848\\
40	1.2114227239884\\
41	1.21142272398835\\
42	1.21142272398833\\
43	1.21142272398832\\
44	1.21142272398832\\
45	1.21142272398831\\
46	1.21142272398831\\
47	1.21142272398831\\
48	1.21142272398831\\
49	1.21142272398831\\
50	1.21142272398831\\
51	1.21142272398831\\
52	1.21142272398831\\
53	1.21142272398831\\
54	1.21142272398831\\
55	1.21142272398831\\
56	1.21142272398831\\
57	1.21142272398831\\
58	1.21142272398831\\
59	1.21142272398831\\
60	1.21142272398831\\
61	1.21142272398831\\
62	1.21142272398831\\
63	1.21142272398831\\
64	1.21142272398831\\
65	1.21142272398831\\
66	1.21142272398831\\
67	1.21142272398831\\
68	1.21142272398831\\
69	1.21142272398831\\
70	1.21142272398831\\
71	1.21142272398831\\
72	1.21142272398831\\
73	1.21142272398831\\
74	1.21142272398831\\
75	1.21142272398831\\
76	1.21142272398831\\
77	1.21142272398831\\
78	1.21142272398831\\
79	1.21142272398831\\
80	1.21142272398831\\
81	1.21142272398831\\
82	1.21142272398831\\
83	1.21142272398831\\
84	1.21142272398831\\
85	1.21142272398831\\
86	1.21142272398831\\
87	1.21142272398831\\
88	1.21142272398831\\
89	1.21142272398831\\
90	1.21142272398831\\
91	1.21142272398831\\
92	1.21142272398831\\
93	1.21142272398831\\
94	1.21142272398831\\
95	1.21142272398831\\
96	1.21142272398831\\
97	1.21142272398831\\
98	1.21142272398831\\
99	1.21142272398831\\
100	1.21142272398831\\
101	1.21142272398831\\
102	1.21142272398831\\
103	1.21142272398831\\
104	1.21142272398831\\
105	1.21142272398831\\
106	1.21142272398831\\
107	1.21142272398831\\
108	1.21142272398831\\
109	1.21142272398831\\
110	1.21142272398831\\
111	1.21142272398831\\
112	1.21142272398831\\
113	1.21142272398831\\
114	1.21142272398831\\
115	1.21142272398831\\
116	1.21142272398831\\
117	1.21142272398831\\
118	1.21142272398831\\
119	1.21142272398831\\
120	1.21142272398831\\
121	1.21142272398831\\
122	1.21142272398831\\
123	1.21142272398831\\
124	1.21142272398831\\
125	1.21142272398831\\
126	1.21142272398831\\
127	1.21142272398831\\
128	1.21142272398831\\
129	1.21142272398831\\
130	1.21142272398831\\
131	1.21142272398831\\
132	1.21142272398831\\
133	1.21142272398831\\
134	1.21142272398831\\
135	1.21142272398831\\
136	1.21142272398831\\
137	1.21142272398831\\
138	1.21142272398831\\
139	1.21142272398831\\
140	1.21142272398831\\
141	1.21142272398831\\
142	1.21142272398831\\
143	1.21142272398831\\
144	1.21142272398831\\
145	1.21142272398831\\
146	1.21142272398831\\
147	1.21142272398831\\
148	1.21142272398831\\
149	1.21142272398831\\
150	1.21142272398831\\
151	1.21142272398831\\
152	1.21142272398831\\
153	1.21142272398831\\
154	1.21142272398831\\
155	1.21142272398831\\
156	1.21142272398831\\
157	1.21142272398831\\
158	1.21142272398831\\
159	1.21142272398831\\
160	1.21142272398831\\
161	1.21142272398831\\
162	1.21142272398831\\
163	1.21142272398831\\
164	1.21142272398831\\
165	1.21142272398831\\
166	1.21142272398831\\
167	1.21142272398831\\
168	1.21142272398831\\
169	1.21142272398831\\
170	1.21142272398831\\
171	1.21142272398831\\
172	1.21142272398831\\
173	1.21142272398831\\
174	1.21142272398831\\
175	1.21142272398831\\
176	1.21142272398831\\
177	1.21142272398831\\
178	1.21142272398831\\
179	1.21142272398831\\
180	1.21142272398831\\
181	1.21142272398831\\
182	1.21142272398831\\
183	1.21142272398831\\
184	1.21142272398831\\
185	1.21142272398831\\
186	1.21142272398831\\
187	1.21142272398831\\
188	1.21142272398831\\
189	1.21142272398831\\
190	1.21142272398831\\
191	1.21142272398831\\
192	1.21142272398831\\
193	1.21142272398831\\
194	1.21142272398831\\
195	1.21142272398831\\
196	1.21142272398831\\
197	1.21142272398831\\
198	1.21142272398831\\
199	1.21142272398831\\
200	1.21142272398831\\
};
\addlegendentry{Prop $(\alpha^{(k)}= 1/2^k)$}

\addplot [color=black, dashed, line width=1.2pt]
  table[row sep=crcr]{%
1	3.0481117080741\\
2	3.02167177446191\\
3	2.99599001942547\\
4	2.9710323445098\\
5	2.94676672624181\\
6	2.92316305798067\\
7	2.9001930061739\\
8	2.87782987949963\\
9	2.85604850955675\\
10	2.83482514192212\\
11	2.81413733653085\\
12	2.79396387645478\\
13	2.77428468425819\\
14	2.75508074520094\\
15	2.73633403663863\\
16	2.71802746303978\\
17	2.70014479610133\\
18	2.68267061949829\\
19	2.66559027785118\\
20	2.64888982953735\\
21	2.63255600301021\\
22	2.61657615632313\\
23	2.60093823958526\\
24	2.58563076010219\\
25	2.57064274997859\\
26	2.55596373598062\\
27	2.54158371147494\\
28	2.5274931102779\\
29	2.51368278226386\\
30	2.50014397059488\\
31	2.48686829044652\\
32	2.4738477091154\\
33	2.46107452740406\\
34	2.4485413621878\\
35	2.43624113007601\\
36	2.4241670320882\\
37	2.41231253927124\\
38	2.40067137919058\\
39	2.38923752323361\\
40	2.37800517466835\\
41	2.36696875740504\\
42	2.35612290541256\\
43	2.34546245274519\\
44	2.3349824241387\\
45	2.32467802613795\\
46	2.3145446387209\\
47	2.30457780738684\\
48	2.29477323567872\\
49	2.28512677811183\\
50	2.27563443348316\\
51	2.26629233853753\\
52	2.25709676196827\\
53	2.2480440987319\\
54	2.23913086465763\\
55	2.23035369133386\\
56	2.22170932125505\\
57	2.21319460321354\\
58	2.2048064879219\\
59	2.1965420238523\\
60	2.18839835328047\\
61	2.18037270852231\\
62	2.17246240835249\\
63	2.1646648545945\\
64	2.15697752887272\\
65	2.14939798951747\\
66	2.14192386861476\\
67	2.13455286919258\\
68	2.12728276253663\\
69	2.12011138562841\\
70	2.11303663869909\\
71	2.10605648289327\\
72	2.09916893803663\\
73	2.09237208050217\\
74	2.08566404117007\\
75	2.07904300347616\\
76	2.07250720154463\\
77	2.06605491840079\\
78	2.05968448425971\\
79	2.05339427488713\\
80	2.04718271002894\\
81	2.04104825190605\\
82	2.03498940377131\\
83	2.02900470852552\\
84	2.0230927473898\\
85	2.01725213863157\\
86	2.01148153634155\\
87	2.00577962925955\\
88	2.0001451396466\\
89	1.99457682220142\\
90	1.9890734630191\\
91	1.98363387859016\\
92	1.97825691483804\\
93	1.97294144619345\\
94	1.96768637470371\\
95	1.96249062917585\\
96	1.95735316435159\\
97	1.95227296011319\\
98	1.94724902071853\\
99	1.94228037406432\\
100	1.93736607097621\\
101	1.93250518452456\\
102	1.92769680936494\\
103	1.92294006110213\\
104	1.91823407567686\\
105	1.91357800877412\\
106	1.90897103525229\\
107	1.9044123485922\\
108	1.89990116036529\\
109	1.89543669972012\\
110	1.89101821288647\\
111	1.88664496269631\\
112	1.88231622812106\\
113	1.87803130382432\\
114	1.87378949972966\\
115	1.86959014060273\\
116	1.86543256564722\\
117	1.86131612811409\\
118	1.85724019492355\\
119	1.85320414629938\\
120	1.84920737541501\\
121	1.84524928805102\\
122	1.84132930226352\\
123	1.83744684806314\\
124	1.83360136710411\\
125	1.8297923123831\\
126	1.82601914794754\\
127	1.82228134861288\\
128	1.81857839968875\\
129	1.81490979671336\\
130	1.81127504519615\\
131	1.80767366036821\\
132	1.80410516694026\\
133	1.8005690988679\\
134	1.79706499912387\\
135	1.79359241947715\\
136	1.79015092027851\\
137	1.78674007025247\\
138	1.78335944629528\\
139	1.7800086332788\\
140	1.77668722386015\\
141	1.77339481829672\\
142	1.77013102426659\\
143	1.76689545669405\\
144	1.76368773758007\\
145	1.76050749583754\\
146	1.75735436713125\\
147	1.75422799372217\\
148	1.75112802431627\\
149	1.74805411391736\\
150	1.74500592368411\\
151	1.74198312079092\\
152	1.73898537829264\\
153	1.73601237499292\\
154	1.73306379531618\\
155	1.73013932918295\\
156	1.72723867188868\\
157	1.7243615239856\\
158	1.72150759116788\\
159	1.71867658415977\\
160	1.71586821860662\\
161	1.71308221496886\\
162	1.71031829841867\\
163	1.70757619873934\\
164	1.70485565022726\\
165	1.7021563915964\\
166	1.69947816588525\\
167	1.69682072036611\\
168	1.69418380645671\\
169	1.69156717963404\\
170	1.68897059935034\\
171	1.68639382895124\\
172	1.68383663559586\\
173	1.68129879017897\\
174	1.67878006725501\\
175	1.67628024496403\\
176	1.67379910495939\\
177	1.67133643233731\\
178	1.66889201556799\\
179	1.66646564642855\\
180	1.66405711993745\\
181	1.6616662342906\\
182	1.65929279079887\\
183	1.65693659382724\\
184	1.6545974507352\\
185	1.65227517181872\\
186	1.64996957025352\\
187	1.64768046203963\\
188	1.64540766594729\\
189	1.64315100346407\\
190	1.64091029874328\\
191	1.63868537855345\\
192	1.63647607222906\\
193	1.63428221162239\\
194	1.63210363105636\\
195	1.62994016727862\\
196	1.62779165941649\\
197	1.625657948933\\
198	1.62353887958396\\
199	1.62143429737584\\
200	1.61934405052477\\
};
\addlegendentry{BP-CLR $(\alpha^{(k)} = 10^{-6})$}

\end{axis}
\end{tikzpicture}%